\documentclass{article}
\usepackage{subfig}
\usepackage{microtype}
\usepackage{graphicx}
\usepackage{booktabs} 
\usepackage{multirow}
\usepackage[utf8]{inputenc}
\usepackage{microtype}
\usepackage{graphicx}
\usepackage{booktabs} 
\usepackage{subfig}
\usepackage{listings}
\usepackage[table]{xcolor}
\usepackage{adjustbox}
\usepackage{tabularx}

\definecolor{lightgreen}{RGB}{200,240,200}
\definecolor{lightred}{RGB}{240,200,200}

\usepackage{amsmath,amsfonts,bm}

\def\eqref#1{equation~\ref{#1}}

\def\1{\bm{1}}

\def\vc{{\bm{c}}}

\def\vr{{\bm{r}}}

\def\mM{{\bm{M}}}

\def\mV{{\bm{V}}}
\def\mW{{\bm{W}}}

\DeclareMathAlphabet{\mathsfit}{\encodingdefault}{\sfdefault}{m}{sl}
\SetMathAlphabet{\mathsfit}{bold}{\encodingdefault}{\sfdefault}{bx}{n}

\newcommand{\R}{\mathbb{R}}

\usepackage{hyperref}

\usepackage{amsmath}
\usepackage{amssymb}
\usepackage{mathtools}
\usepackage{amsthm}
\usepackage{pifont}
\usepackage[capitalize,noabbrev]{cleveref}

\theoremstyle{plain}

\theoremstyle{definition}

\theoremstyle{remark}

\usepackage{booktabs}
\usepackage{multirow}
\usepackage{tikz}
\usepackage{ifthen}
\newcommand{\cmark}{\ding{51}} 
\newcommand{\xmark}{\ding{55}} 
\usepackage[textsize=tiny]{todonotes}
\usepackage{hyperref}
\usepackage{listings}
\usepackage{xcolor}
\usepackage{xcolor}
\newcommand{\rev}[1]{{#1}}
\usepackage[accepted]{mlsys/mlsys2025}
\newcommand{\addartifactbadges}{%
  \begin{tikzpicture}[overlay, remember picture]
    \node[xshift=-1.5cm, yshift=-1.1cm] at (current page.north east)
      {\includegraphics[width=1.9cm]{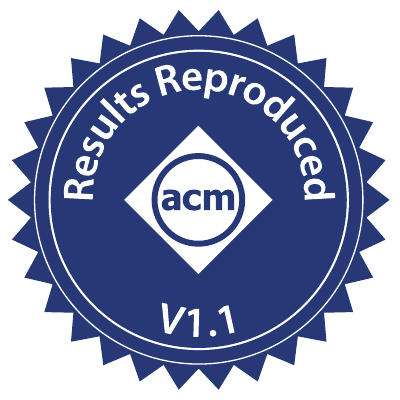}};
    
    \node[xshift=-3.8cm, yshift=-1.1cm] at (current page.north east)
      {\includegraphics[width=1.9cm]{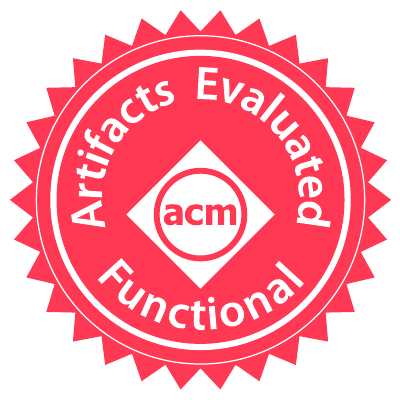}};
    
    \node[xshift=-6.1cm, yshift=-1.1cm] at (current page.north east)
      {\includegraphics[width=1.9cm]{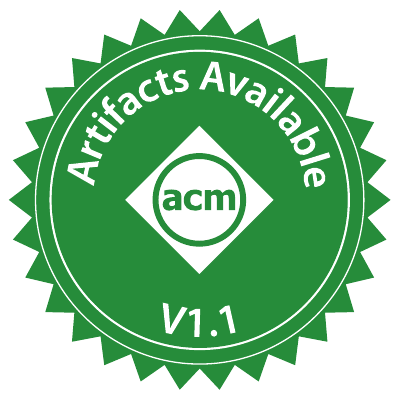}};
  \end{tikzpicture}%
}

\mlsystitlerunning{PyLO: Towards Accessible Learned Optimizers in PyTorch
}

\begin{document}

\twocolumn[

\mlsystitle{PyLO: Towards Accessible Learned Optimizers in PyTorch}



\mlsyssetsymbol{equal}{*}

\begin{mlsysauthorlist}
\mlsysauthor{Paul Janson}{equal,con,mila}
\mlsysauthor{Benjamin Thérien}{equal,udem,mila}
\mlsysauthor{Quentin Anthony}{elu}
\mlsysauthor{Xiaolong Huang}{con,mila}
\mlsysauthor{ Abhinav Moudgil}{con,mila}
\mlsysauthor{Eugene Belilovsky}{con,mila}

\end{mlsysauthorlist}

\mlsysaffiliation{con}{Concordia University}
\mlsysaffiliation{mila}{Mila Quebec AI Institute}
\mlsysaffiliation{udem}{Université de Montréal}
\mlsysaffiliation{elu}{Eluther AI}

\mlsyscorrespondingauthor{Paul Janson}{paul.janson@mila.quebec}
\mlsyscorrespondingauthor{Benjamin Thérien}{benjamin.therien@mila.quebec}

\mlsyskeywords{Machine Learning}

\vskip 0.3in

\begin{abstract}
Learned optimizers have been an active research topic over the past decade, with increasing progress toward practical, general-purpose optimizers that can serve as drop-in replacements for widely used methods like Adam. However, recent advances such as VeLO, which was meta-trained for 4000 TPU-months, remain largely inaccessible to the broader community, in part due to their reliance on JAX and the absence of user-friendly packages for independently using the optimizers after meta-training. To address this gap, we introduce PyLO, a PyTorch-based library that brings learned optimizers to the remaining \rev{$\approx 70\%$} of machine learning community \rev{~\cite{linuxfoundation2024annual}} via the familiar \emph{torch.optim.Optimizer} interface. Unlike prior work focused on limited-scale academic tasks, our emphasis is on applying learned optimization to real-world large-scale pre-training tasks. Our systems contribution includes CUDA-accelerated implementations of the \texttt{small\_fc\_lopt}\cite{metz2022practical} and VeLO\cite{velo} learned optimizers, achieving substantial performance gains, with training throughput on ViT-B/16 (batch size 32) increasing from 39.36 and 49.73 to 205.59 and 191.18 samples per second, respectively. PyLO has the versatility that allows us to easily combine learned optimizers with existing optimization tools, such as learning rate schedules and weight decay. When doing so, we discover that learned optimizers can substantially benefit from it. Our code is available at
\href{https://github.com/Belilovsky-Lab/pylo}{https://github.com/Belilovsky-Lab/pylo .}

\end{abstract}
]
\addartifactbadges
\printAffiliationsAndNotice{\mlsysEqualContribution}

\begin{figure}[h]
    \centering
    \includegraphics[width=1.01\linewidth]{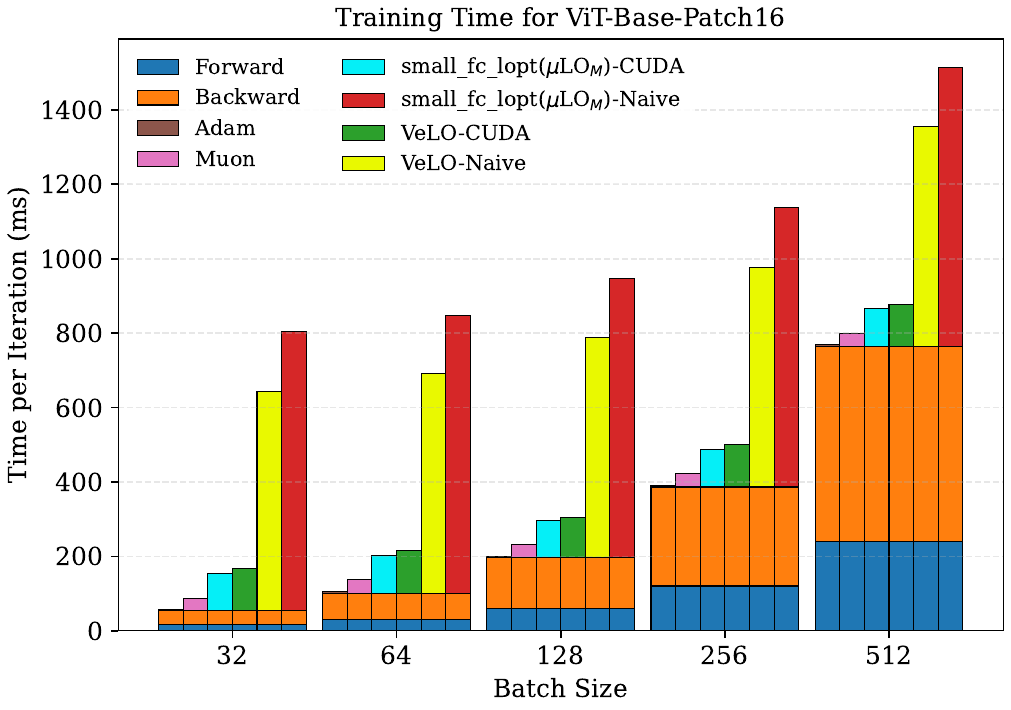}
    \caption{\textbf{Training step timing breakdown for ViT-B/16 on a single A100 GPU.} 
             We measure the 
             \textcolor{blue}{forward}, 
             \textcolor{orange}{backward}, and 
           optimizer 
             times per training step. 
             We observe that the CUDA-accelerated learned optimizer steps 
             (\textcolor{cyan}{cyan}, \textcolor{green}{green}) 
             show substantial improvements over the naive implementations 
             (\textcolor{yellow}{yellow}, \textcolor{red}{red}). 
             In all cases, as the batch size is increased, the relative overhead of the optimizer shrinks.}
    \vspace{-16pt}
    \label{fig:vit-b16_1gpu}
\end{figure}

\section{Introduction}

\begin{figure*}[t]
    \centering
    \includegraphics[width=0.9\linewidth]{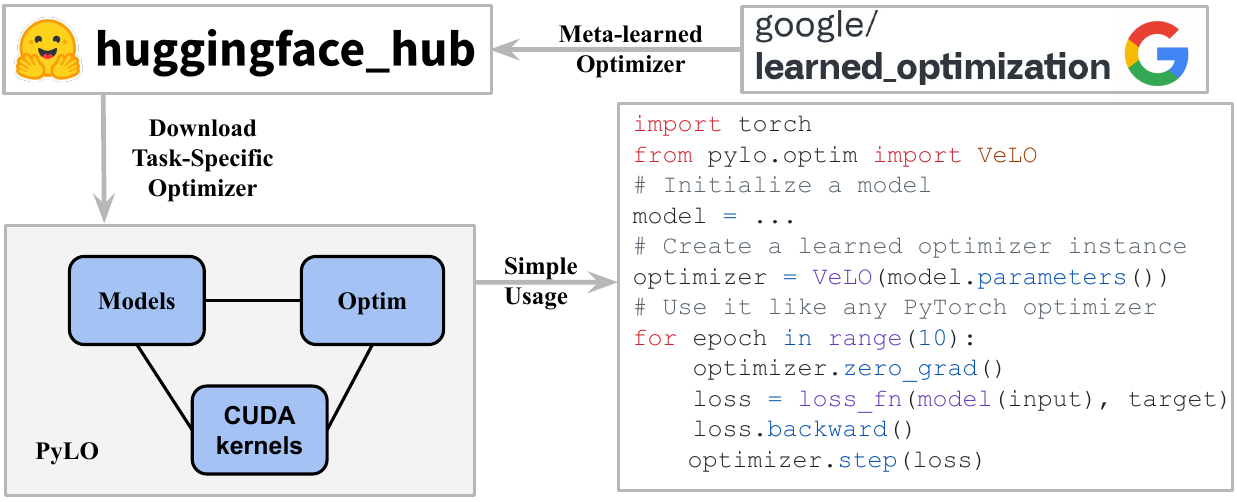}\vspace{-10pt}
    \caption{\textbf{PyLO:} simplifies the integration of learned optimizers into standard machine learning workflows. By addressing key usability challenges, PyLO provides seamless access to meta-learned optimization techniques through three core features: (1) automatic weight loading from Hugging Face Hub, (2) a familiar PyTorch-style optimizer interface, and (3) accelerated CUDA kernel support. The library bridges the gap between advanced meta-learning research (\texttt{google/learned\_optimization}\cite{metz2022practical}) and practical machine-learning applications, enabling researchers and practitioners to easily leverage state-of-the-art learned optimization techniques in PyTorch.}\vspace{-10pt}
    \label{fig:pylo-overview}
\end{figure*}

Learned optimization (LO)~\citep{hochreiter2001learning,andrychowicz2016learning} represents a promising yet underexplored direction for advancing optimization algorithms in machine learning. Training contemporary neural networks requires solving highly non-convex optimization problems for which theory neither guarantees convergence to a globally optimal solution, nor convergence at the optimal rate. Therefore, despite the purportedly strong performance of hand-designed optimizers such as Adam~\citep{kingma2017adam}, for many tasks of interest, there may be substantial room for improvement through learning to optimize. Indeed, VeLO~\cite{velo}, the most performant publicly available learned optimizer to date, was shown to outperform a well-tuned NadamW and schedule baseline optimizer without hyperparameter tuning. Despite its strong performance, however,  VeLO is seldom used by our community today.

 Several critical barriers have hindered the more widespread adoption of learned optimizers for deep learning applications: 1) an overfocus on meta-learning in current learned optimization frameworks, 2) the absence of a PyTorch implementation for the most performant learned optimizers available, 3) the inability to effectively share optimizer weights, and 4) learned optimizer step overhead. 

In what follows, we take a step towards improving the adoption of learned optimizers for deep learning by providing a PyTorch implementation of state-of-the-art learned optimizers that emphasizes their performance and accessibility for practical machine learning problems. our contributions can be summarized as follows:
\vspace{-5pt}
\begin{enumerate}
  \setlength\itemsep{0em}
    \item We provide a modular open-source implementation of state-of-the-art learned optimizers in PyTorch, which seamlessly integrates with \texttt{torch.optim.Optimizer} and the Huggingface ecosystem to easily integrate with existing code and facilitate standardized sharing of task-specific learned optimizer weights (\autoref{fig:pylo-overview}).
    \item We provide a CUDA-accelerated implementation of \texttt{small\_fc\_lopt} and VeLO's optimizer step, which substantially reduces memory overhead and increases occupancy (GPU utilization as in \autoref{fig:vit-b16_1gpu}).  
    \item We benchmark optimizer step times and performance in PyLO for popular image classification and language modeling workloads, showing that our implementation scales to larger workloads and decreases step times by more than 2$\times$ over the existing JAX implementation. 
    \item We demonstrate that learned optimizer step times can further be reduced in a data-parallel training setting by distributing the optimizer step across devices.
    \item Our flexible framework makes it easy to study the effect of weight decay and learning rate schedules on learned optimizer performance by leveraging the existing Pytorch API and, when doing so, we find that making these simple additions can substantially improve performance in some cases. 
\end{enumerate}
\vspace{-5pt}

\section{Challenges of L2O adoption for the Wider deep learning community}
This section reviews existing open-source learned optimization repositories and reflects on the critical barriers hindering the widespread adoption of learned optimizers within the broader deep learning community, highlighting pitfalls in existing work.

\begin{table*}[h]
\centering
\caption{\textbf{Comparison of Open-source Learned Optimization Repositories.} We list the current open-source repositories for learned optimization and list their status relative to the difficulties of L2O adoption from section~\ref{sec:adoption}. A Decoupled framework separates learned optimizer implementation from meta-training and meta-testing code. }
\label{tab:learned_opt_comparison}
\begin{adjustbox}{width=\textwidth,center}
\begin{tabular}{l|cccc|cc}
\toprule
\textbf{Repository} & \textbf{Decoupled} & \textbf{PyTorch} & \textbf{HuggingFace} & \textbf{CUDA Kernel} & \textbf{Paper} & \textbf{Github} \\
\midrule
\textbf{Open-L2O} & \xmark & \cmark &  \xmark & \xmark & \cite{chen2022learning} & \href{https://github.com/VITA-Group/Open-L2O}{Open-L2O} \\
\textbf{Learned Optimization} & \xmark & \xmark & \xmark &  \xmark & \cite{metz2022practical} & \href{https://github.com/google/learned_optimization}{Learned Optimization} \\
\textbf{PyLO (ours)} & \cmark & \cmark & \cmark &\cmark & \textit{In submission} & \href{https://anonymous.4open.science/r/pylo-C91E}{PyLO}\\

\bottomrule
\end{tabular}
\end{adjustbox}
\end{table*}

\subsection{Existing Open source libraries}

\textbf{Open-L2O} \cite{chen2022learning} offers a comprehensive benchmarking suite designed primarily for evaluating learned optimizers (L2O) across a range of different optimization problems: convex functions, non-convex functions, minimax functions, and some neural network training. It includes both model-based and model-free L2O methods, facilitating reproducibility and fair comparisons, but only includes a limited evaluation of the learned optimizer for practical deep learning tasks. In contrast, PyLO's main focus is on learned optimization for deep learning.

\textbf{Google's learned\_optimization} \cite{metz2022practical} is a research-oriented repository written in JAX for training, designing, evaluating, and applying learned optimizers. It supports an extensive set of features for meta-training optimizers: many gradient estimation algorithms, abstractions for different types of tasks, various task modifiers, etc. While the repository also provides functionality for applying pre-trained optimizers to new tasks via the Optax interface~\cite{deepmind2020jax}, its main focus is clearly to provide meta-training functionality.

\subsection{Challenges to adoptions}
\label{sec:adoption}

\textbf{Overfocus on meta-learning:}
Existing libraries, such as the Learned Optimization~\citep{metz2022practical,velo} and Open L2O~\cite{chen2022learning}, prioritize meta-training functionality over precisely focusing on the practical deployment of learned optimizers. This leads to a large proportion of the code being unnecessary for applying learned optimizers to new tasks. This can represent a barrier to entry, particularly for researchers and practitioners who seek to leverage pre-trained optimization algorithms without engaging in the complexities of meta-training (similar to using a pretrained transformer from \texttt{huggingface/transformers\cite{wolf-etal-2020-transformers}}). This highlights the necessity of decoupling learned optimizer implementations from their meta-training and meta-testing code.

\textbf{PyTorch v.s. JAX:} A third barrier relates to the underlying deep learning framework. While JAX~\citep{jax2018github} offers powerful functional programming capabilities and strong performance from compilation, it lacks the widespread community adoption of PyTorch~\citep{pytorch}. Beyond having more users, widespread community adoption also comes with a more mature and feature-full open source ecosystem~\cite{he2019mlframeworks,rw2019timm,von-platen-etal-2022-diffusers,wolf-etal-2020-transformers}. With so much demand for compatibility with PyTorch,  there exists a pressing need for a modular L2O framework that seamlessly integrates with PyTorch's optimizer interface. Currently, Google's Learned Optimization does not provide any compatibility with PyTorch. While Open-L2O does support PyTorch, it lacks many other crucial aspects for adoption (see~\autoref{tab:learned_opt_comparison}).

\textbf{Sharing learned optimizer weights:} Being able to effectively share optimizer weights and document their training distribution and evaluation performance is essential for building an effective open source ecosystem around learned optimizers. However, no such mechanism currently exists. In contrast, pre-trained models for language understanding~\citep{deepseekr1,llama2}, image classification~\citep{dosovitskiy2020image,clip}, semantic segmentation~\citep{segment_anything}, and a number of other tasks are routinely shared via platforms such as HuggingFace Hub~\citep{huggingface_hub}, enabling practitioners to leverage computational investments made by well-resourced institutions. This collaborative open-model ecosystem has accelerated innovation and reduced redundancy in these domains through its expansion of access to state-of-the-art models. Despite this revolution in model sharing, we lack seamless sharing in learned optimizer weights. We believe that Learned Optimizer (LO) models can and should be integrated into the community through the HuggingFace ecosystem. Current open-source L2O frameworks, Learned Optimization and OpenL2O, do not provide any mechanism for sharing learned optimizer weights.

\textbf{Per-step overhead of learned optimizers:} A significant obstacle to the widespread adoption of learned optimizers is the computational overhead they can introduce to already resource-intensive training setups. Deep learning workflows operate under strict time and computational constraints, with practitioners demonstrating marked reluctance to trying new optimization strategies that might increase training durations, regardless of potential convergence benefits. This challenge necessitates meticulous implementation of learned optimization algorithms to ensure competitive computational efficiency, which can be achieved by writing custom CUDA kernels.

\section{Project goal}
With our goal of improving the adoption of learned optimizers within the deep learning community and enhancing the open-source ecosystem around them, we design PyLO with the following principles in mind:

\textbf{Accessibility} With minimal dependencies (PyTorch + HuggingFace) and a straightforward installation process, PyLO eliminates the technical barriers that have impeded the widespread adoption of learned optimizers thus far.

\textbf{Decoupling} By exclusively providing efficient learned optimizer implementations in PyTorch without meta-training code, PyLO decouples meta-testing from meta-training. This substantially reduces code bloat and additional dependencies required compared to other frameworks. As a result, PyLO is simpler to use and easier to understand.

\textbf{Interoperability} Through strict adherence to the PyTorch optimizer (\textit{torch.optim.Optimizer}) interface, PyLO ensures seamless integration with existing training frameworks and compatibility with popular optimizer modifiers. This allows practitioners to use learned optimizers with minimal modification to their existing code---including modifiers like decoupled weight decay and learning rate schedules.

\section{Library design}

We implement a modular architecture partitioned into four main components to realize our project vision. This design philosophy prioritizes both usability for practitioners and extensibility for researchers developing novel learned optimizers.\\

\textbf{Optimization Module} (\textit{pylo.optim}) This module facilitates critical state management functions necessary for learned optimization. These include maintaining parameter-specific accumulators, computing parameter features for optimizer forward passes, executing update steps with configurable learning rate scaling, and supporting standard PyTorch optimizer functionality such as state dictionaries for resuming. \rev{It also composes well with torch.compile and activation checkpointing}. This implementation allows practitioners to leverage advanced learned optimization techniques without significant modifications to existing training setups. \\

\textbf{Meta-Model Architectures} \textit{(pylo.models)} encapsulates the parameters of the learned optimizer and its forward pass.  It is also fully integrated with HuggingFaceHub~\citep{huggingface_hub}, allowing users to download existing optimizers from the hub and providing a mechanism for weight distribution and versioning of future optimizers. Through HF integration, we facilitate community-driven improvement cycles and allow researchers to easily leverage our CUDA-accelerated implementation for their own benchmarking.

\textbf{CUDA Acceleration} (\textit{pylo.csrc}) The computational demands of learned optimizers present significant implementation challenges. Unlike traditional optimization algorithms that apply simple, closed-form update rules, learned optimizers require evaluating a small multilayer perceptron (MLP) for each parameter in the optimizee. This can introduce substantial computational overhead for large models, creating a critical bottleneck that limits practical deployment.
We emphasize that standard PyTorch neural network modules are inadequately optimized for this specific use case. These modules are primarily designed for batched operations on image or language data rather than the unique computational pattern of learned optimizers, where many small MLPs must be evaluated in parallel across millions or billions of parameters.

Drawing inspiration from architecture-aware optimizations~\citep{muller2021real, dao2022flashattention}, we implemented specialized CUDA kernels for applying learned optimizers. This implementation strategically leverages the GPU memory hierarchy to address the primary bottleneck: memory bandwidth rather than computational throughput.
Our CUDA implementation employs two key optimization strategies: (1) \textbf{Fused MLP Inference} and (2) \textbf{On-demand Feature Computation}. The first utilizes GPU registers to store intermediate activations during forward propagation through the optimizer's MLP. This approach minimizes high-latency global memory accesses by keeping frequently accessed data in the fastest available memory tier. The second avoids storing normalized features in global memory, recalculating them when needed, and trading redundant computation for reduced memory bandwidth. The full implementation and performance analysis are detailed in Section~\ref{sec:cuda-impl}.

\paragraph{Decoupled evaluation} Providing easy-to-use examples for new open-source code can go a long way for improving adoption, as it illustrates how the library can be used and provides a starting point for exploration. However, it is important to decouple such examples from the learned optimizer implementation itself to keep the lines of code within PyLO to a minimum. As such, we provide 
\href{https://github.com/Belilovsky-Lab/pylo_examples}{PyLO-Examples} 
alongside PyLO as a supporting repository for evaluating learned optimizers on popular language modelling (pre-training on FineWeb EDU) and image classification (classical ImageNet pre-training) tasks. Each task is implemented within a single directory to maximize readability and re-use.

\section{CUDA Implementation} 
\label{sec:cuda-impl}
In this section, we describe the implementation details of our CUDA kernels and the motivation behind our design choices. We begin by analyzing the performance bottlenecks in the naive PyTorch implementation, then present our optimized approach.

\subsection{Analysis of Naive Implementation}

\begin{figure*}[t]
    \centering
    \includegraphics[width=1.0\linewidth]{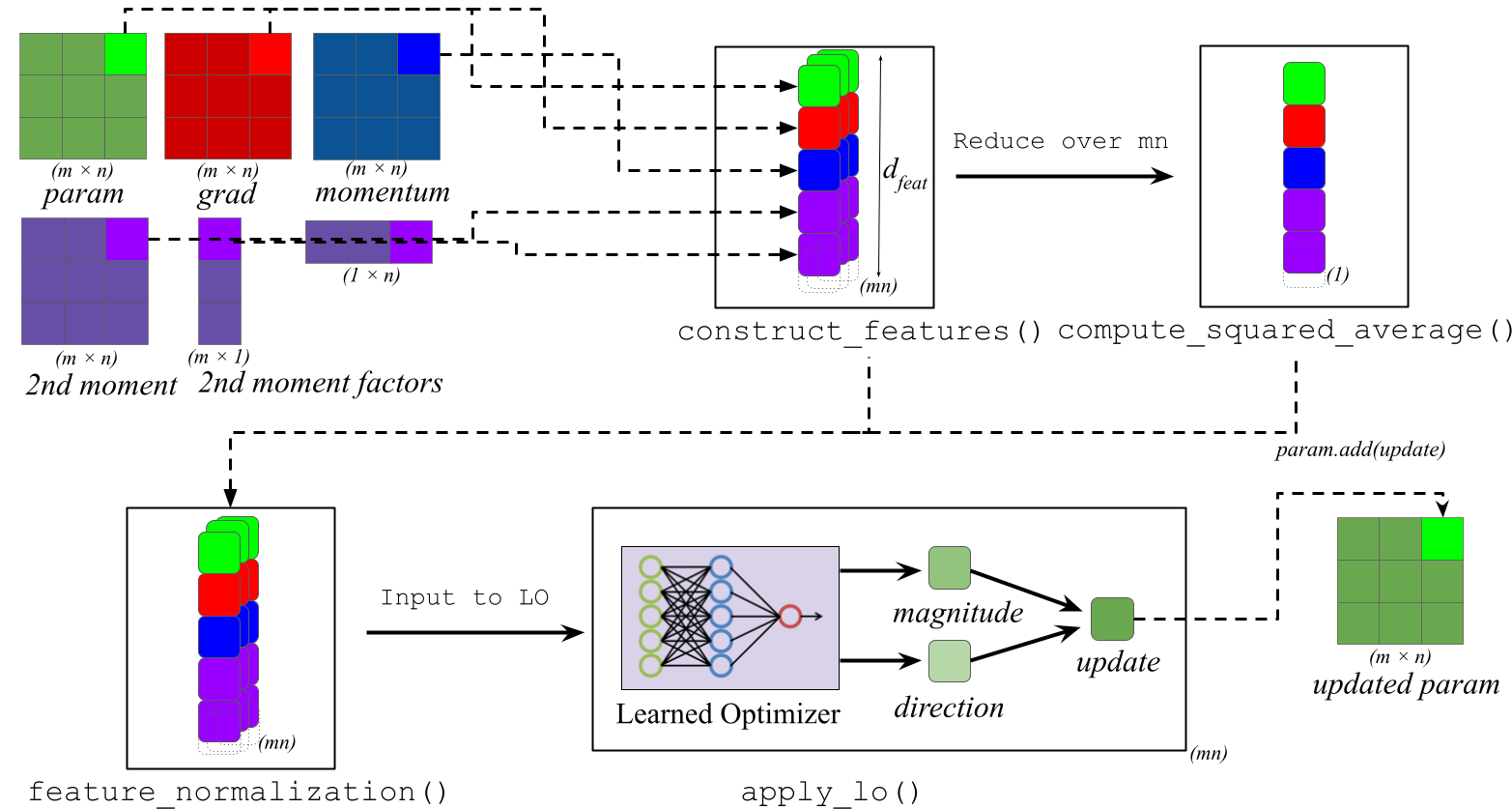}
    \vspace{-5mm}
    \caption{\textbf{Overview of the Learned Optimizer (LO) update mechanism:} The figure depicts the computation pipeline used by learned optimizers such as \texttt{small\_fc\_lopt} and VeLO. Model parameters (\(\mathbf{\theta}\)), gradients (\(\mathbf{g}\)), momentum (\textit{m}), and second-moment accumulators (\textit{v}, row \& column factors of \textit{v}) are combined within \texttt{construct\_features()} to form a rich set of derivative features, including elementwise interactions such as \(\mathbf{g} \times \text{momentum}\) and \(\text{momentum} \times \text{factors}\) (see Appendix \ref{apdx:features_learned_opt} for a complete description). These features are reduced over the parameter dimensions (\(m \times n\)) in \texttt{compute\_squared\_average()} to obtain normalization statistics. In \texttt{feature\_normalization()}, features are normalized before being processed by the Learned Optimizer, which predicts an update direction and magnitude in \texttt{apply\_lo()}, and this predicted update is then applied to yield the updated parameters.}
    \label{fig:computation_pipeline}
\end{figure*}

A learned optimizer step comprises several distinct operations, visualized in Figure \ref{fig:computation_pipeline}. Consider a parameter tensor of dimensions $m \times n$. The naive implementation constructs a feature vector of size $mn \times d_{\text{feat}}$ where $d_{\text{feat}}$ denotes the learned optimizer's input dimensionality. This construction starts after the accumulator update (update of momentum, second moment, etc) and proceeds in three stages:


\textbf{Feature Construction} After updating accumulators the \texttt{construct\_features()} procedure starts. The gradient $g$, parameter $p$, momentum buffer $m$, second-moment estimate $v$, and factored second-moment terms (row and column factors), are used to compute a set of derived features (see Appendix \ref{apdx:features_learned_opt} for all features). Each derived feature requires individual kernel launches for element-wise operations.

\textbf{Feature normalization} Before applying the learned optimizer's neural network, features must be normalized by their squared average across all parameter elements. This requires: (1) computing $\mathbb{E}[\text{feature}^2]$ via reduction operations (\texttt{compute\_squared\_average()}), and (2) dividing each feature by $\sqrt{\mathbb{E}[\text{feature}^2] + \epsilon}$ (\texttt{feature\_normalization()}). PyTorch's general-purpose reduction kernels require additional kernel launches after materializing the full feature vector.

\textbf{Apply Learned Optimizer}  
Following feature preparation and normalization, the \texttt{apply\_lo()} procedure is invoked. A learned optimizer, which is implemented as a neural network, takes the normalized feature vector as input and predicts two scalar outputs: one representing the \emph{magnitude} and the other the \emph{direction} of the parameter update. These predictions are combined to compute the update according to the formula
\[
\Delta \theta = \text{direction} \times e^{\text{magnitude} \times \alpha} \times \beta,
\]
where $\alpha$ and $\beta$ are hyperparameters fixed at $0.01$. The parameter $\theta$ is then updated as $\theta \leftarrow \theta + \Delta \theta$. In the PyTorch-based MLP implementation, this process launches a separate kernel for each layer and stores intermediate activations in global memory. Consequently, memory bandwidth emerges as the primary bottleneck, resulting in underutilized GPU compute resources and reduced overall efficiency. 

\begin{figure*}[t]
    \centering
    \includegraphics[width=1.0\textwidth]{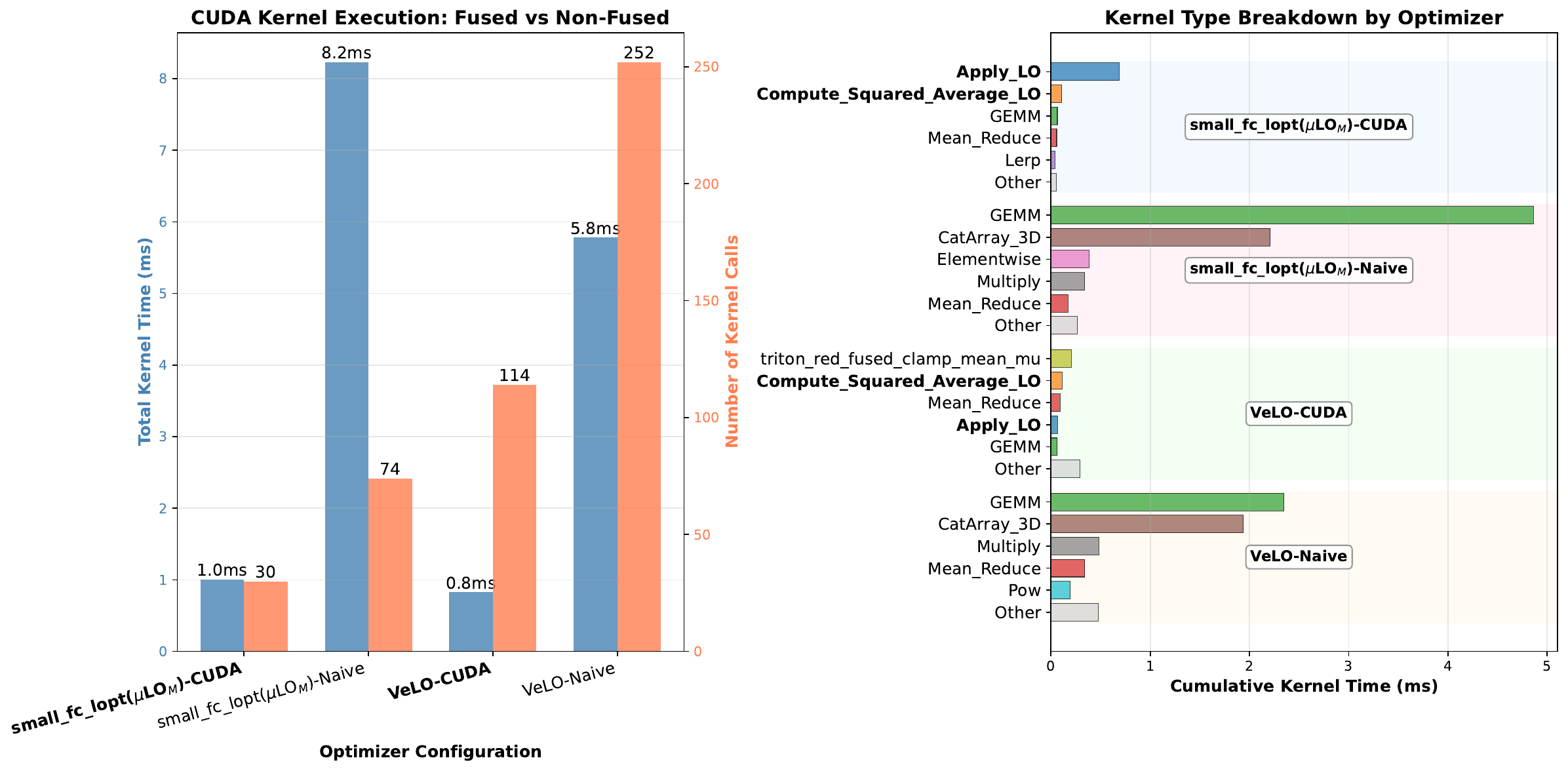}
    \caption{Comparison of CUDA kernel execution characteristics for the learned optimizers \texttt{small\_fc\_lopt} and VeLO, shown for both na\"ive and fused implementations applied to a single layer MLP (shape [1000x1000]) with no bias. \rev{Note that kernel fusion here refers exclusively to fusing the learned optimizer's feature construction, normalization, and MLP inference — the optimizee's forward and backward passes remain unchanged.} (Left) The total cumulative kernel execution time (\textcolor{blue}{blue}) and the number of kernel launches (\textcolor{orange}{orange}) reveal that the na\"ive implementations spend substantially more time executing a far greater number of small kernels, due to performing each arithmetic or reduction step as an independent operation. In contrast, the fused CUDA versions combine these operations into fewer, more computationally dense kernels, reducing both launch overhead and memory bandwidth pressure. (Right) Breakdown of cumulative kernel time by kernel type for each optimizer shows that the fused versions \textbf{(bolded)} consolidate many repetitive operations into specialized fused kernels. This pattern is consistent across both VeLO and \texttt{small\_fc\_lopt}, demonstrating that kernel fusion is broadly beneficial for improving learned optimizer efficiency. \rev{Descriptions about the kernel names provided in Appendix \ref{apdx:kernel_names}}}
    \label{fig:profiler}
\end{figure*}

The naive implementation thus requires 74-252 kernel launches per parameter tensor, with each intermediate result materialized in global memory. For a parameter tensor of size $mn$, this generates approximately $(num\_of\_accum) * mn$ words of memory traffic and allocates $mn \times d_{\text{feat}}$ words of temporary storage, which a substantial burden for large models. This is evidenced by the profiler visualization as shown in Figure \ref{fig:profiler}.

\subsection{Optimized Kernel Design}

Our optimization strategy rests on two key observations: (1) feature construction can be performed in a single pass over the parameter tensor, and (2) the neural network forward pass can be fused with feature construction to eliminate intermediate storage. To this end, we implement a two-pass method in which we decompose the learned optimizer step into two kernels:

\textbf{Kernel 1: Feature statistics collection.} We perform a single pass over all parameters, computing features on-the-fly in registers and accumulating their squared values for normalization. Rather than materializing the full feature tensor, we maintain only a $d_{\text{feat}}$-dimensional accumulator per thread. Reduction proceeds hierarchically: thread-local accumulators are reduced within each warp using shuffle instructions, warp results are aggregated via shared memory, and block-level results are combined through atomic operations to global memory. This approach computes the required statistics using only $O(d_{\text{feat}})$ shared memory per block and $O(d_{\text{feat}})$ global memory total, independent of parameter count. This combines the \texttt{construct\_features()} and \texttt{compute\_squared\_average()} procedures. \\
\textbf{Pseudocode for Kernel 1}
{\small
\begin{lstlisting}
// Phase 1: Fused Compute Statistics
Kernel_1(g, p, m, v, ...) {
    thread_accum[d_feat] = {0};  // Register array
    for (i in grid_stride_loop) {
        features = compute_features(g[i], p[i], m[i], v[i], ...);
        thread_accum += features * features;
    }
    warp_reduce(thread_accum);  // Shuffle operations
    block_reduce(warp_results);  // Shared memory
    atomic_add(global_stats, block_result);
}
\end{lstlisting}
}

\textbf{Kernel 2: Second Fused feature construction and network application.} With normalized statistics available in global memory, we perform a second pass that: (1) loads the normalization factors into shared memory for efficient broadcast, (2) constructs features in registers from parameter and accumulator values for the second time, (3) evaluates the learned optimizer network in-line, and (4) applies the computed update. No intermediate tensors are materialized—all computation flows through the register file. This combines \texttt{construct\_features()},\texttt{feature\_normalization()} and \texttt{apply\_lo()} procedures.

\textbf{Memory hierarchy exploitation} Our design carefully manages data placement across the GPU memory hierarchy:

\textbf{Registers:} Feature vectors ($d_{\text{feat}} \approx 39$ elements \rev{for \texttt{small\_fc\_lopt} and $d_{\text{feat}} \approx 30$ elements for VeLO}) and network activations reside entirely in registers during computation, providing single-cycle access latency. \rev{Our kernels are optimized for the current feature dimensions, but the feature dimension can be changed by tuning kernel launch parameters, with advantages maintained up to the SRAM capacity per streaming multiprocessor. We note that these feature sets have been heavily ablated in prior work~\cite{metz2022practical}, with additional features showing diminishing returns.}

\textbf{Shared memory:} The $d_{\text{feat}}$-dimensional normalization statistics are loaded once per thread block and broadcast to all threads, amortizing the cost of global memory access across all threads in the block.

\textbf{Global memory:} Only parameter values, gradients, and optimizer state buffers are read from global memory, and only updated parameters are written back. This reduces memory traffic substantially per optimization step.

This design eliminates kernel launch overhead (reducing from 74-252 to 30-114 kernel invocations, see Figure \ref{fig:profiler}), removes intermediate allocations (reducing memory footprint by $mn \times d_{\text{feat}}$ words), and minimizes memory bandwidth consumption (Data movement is minimized as only accumulators are read and final outputs are written). The result is an implementation that achieves good memory bandwidth utilization while maintaining numerical equivalence to the naive approach.\\

\textbf{Pseudocode for Kernel 2}
{\small
\begin{lstlisting}
// Phase 2: Fused Apply
Kernel_2(g, p, m, v, ..., global_stats) {
    __shared__ normalized_stats[d_feat];
    if (tid < d_feat) {
        normalized_stats[tid] = rsqrt(global_stats[tid] / N); // N = number of params
    }
    __syncthreads();
    
    for (i in grid_stride_loop) {
        // All computation in registers
        features[d_feat] = compute_features(g[i], p[i], m[i], v[i], ...);
        features *= normalized_stats;  // Broadcast from SMEM
        
        // Inline MLP (weights in __ldg cache)
        hidden = relu(Linear1(features));
        hidden = relu(Linear2(hidden));
        direction, magnitude = Linear3(hidden);
        
        update = direction * exp(magnitude * alpha) * beta;
        p[i] -=  update;
    }
}
\end{lstlisting}
}

\section{Benchmarking LO step times in PyLO}

In this section, we establish the per-step overhead of learned optimizers for common pre-training workloads and showcase the improved scaling behavior of PyLO's CUDA-accelerated learned optimizer steps. Our goal is to report learned optimizer overhead relative to hand-designed optimizers for practical tasks and to illustrate how the overhead scales as the model size is increased. To this end, we benchmark and record the forward, backward, and optimizer step times for training vision transformers and transformer language models of different sizes on a single $80$GB A100 GPU. \\
\textbf{Learned Optimizer overhead shrinks as batch size increases.} Figure~\ref{fig:vit-b16_1gpu} reports timings at different batch sizes for ViT-B/16. We observe that as batch size is increased, the overhead of the learned optimizer's step relative to the forward and backward passes diminishes. This indicates that large-scale training with prolonged per-step compute may be particularly well-suited for learned optimizers, as the optimizer cost can be effectively amortized. This trend is further evidenced in Table~\ref{tab:timing-main-paper} for both vision and language modeling, which presents timing comparisons across optimizers for ViT-B/16 and a medium-sized GPT-2 model (355M parameters). We find that relative throughput compared to Adam improves with model scale, holding true for language modeling as well. Additionally, per-step time for \texttt{small\_fc\_lopt} (CUDA) scales with model size, yet our CUDA kernels deliver substantial speedups over the naive baseline. For ViT-B/16, \texttt{small\_fc\_lopt} (CUDA) achieves an 86\% reduction in optimizer step time versus the naive version; for the GPT-2 task, the reduction is 88\%. (Full results, including larger ViT and GPT variants, are provided in Appendix~\ref{apdx:full-timing}.)

\begin{figure}[t]
    \centering
    \includegraphics[width=1.0\linewidth]{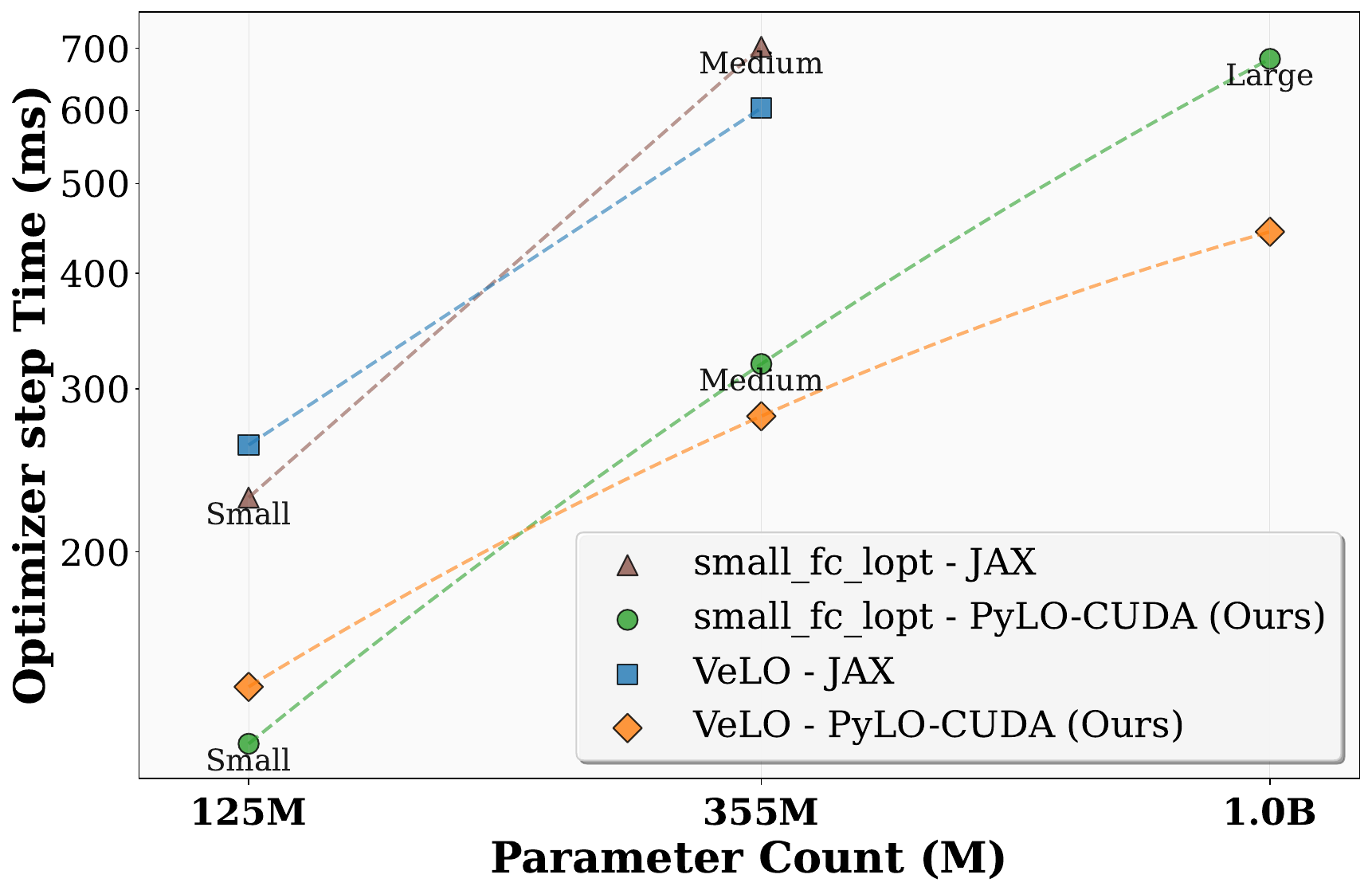}\vspace{-5pt}
    \caption{\textbf{Step Time Scaling of Learned Optimizers.} 
We present a comparison of optimizer step time between our custom CUDA implementation and the original JAX versions of \texttt{small\_fc\_lopt} and VeLO, evaluated during the training of GPT-2 style transformer models across a range of model sizes. The results show that our CUDA implementation not only achieves substantially lower step times but also maintains this advantage as model size increases, enabling more efficient scaling to larger architectures.
}\vspace{-6pt}
    \label{fig:scaling-timing}
\end{figure}

\begin{table}[t]
\centering
\resizebox{1.0\linewidth}{!}{%
\begin{tabular}{@{}ccccccc@{}}
\toprule
\textbf{Model}              & \textbf{Optimizer}      & \textbf{BS/SL} & \textbf{Fwd (ms)} & \textbf{Bwd (ms)} & \textbf{Opt step (ms)} & \textbf{\#Params (M)} \\ \midrule
\multirow{6}{*}{ViT-B/16}   & Adam\cite{adamw}                   & \multirow{6}{*}{32/197} & \multirow{6}{*}{17.52} & \multirow{6}{*}{38.13} & 4.90   & \multirow{6}{*}{86.57} \\
                            & Adafactor\cite{shazeer2018adafac}               &                        &                        &                        & 18.99  &                        \\
                            & \texttt{small\_fc\_lopt} (naive) &                        &                        &                        & 756.80 &                        \\
                            & \texttt{small\_fc\_lopt} (CUDA)  &                        &                        &                        & 99.59 (\textcolor{red}{-86\%})  &                        \\
                            & VeLO (naive)            &                        &                        &                        & 585.11 &                        \\
                            & VeLO (CUDA)             &                        &                        &                        & 113.58 (\textcolor{red}{-80\%}) &                        \\ \midrule
\multirow{6}{*}{GPT2-355M}  & Adam\cite{adamw}                   & \multirow{6}{*}{4/1024} & \multirow{6}{*}{197.41} & \multirow{6}{*}{392.29} & 20.12   & \multirow{6}{*}{355.92} \\
                            & Adafactor\cite{shazeer2018adafac}              &                        &                        &                        & 35.11   &                         \\
                            & \texttt{small\_fc\_lopt} (naive) &                        &                        &                        & 2872.17 &                         \\
                            & \texttt{small\_fc\_lopt} (CUDA)  &                        &                        &                        & 319.14 (\textcolor{red}{-88\%})  &                         \\
                            & VeLO (naive)            &                        &                        &                        & 2378.93 &                         \\
                            & VeLO (CUDA)             &                        &                        &                        & 284.37 (\textcolor{red}{-88\%})  &                         \\ \bottomrule
\end{tabular}%
}
\caption{\textbf{Timing Results Across Optimizers for ViT and GPT Models.} 
We report optimizer step times for Vision Transformer (ViT-B/16) and a GPT-2 style model with 355M parameters, using realistic batch sizes (BS) and sequence lengths (SL) that fit within the memory constraints of a single A100 GPU. The results demonstrate that our custom CUDA kernel significantly reduce optimizer step time \textcolor{red}{(reduction shown in red)} compared to naive implementation, enabling more efficient training in practical settings.
}

\label{tab:timing-main-paper}
\end{table}

\begin{figure*}[t]
    \centering
    \subfloat[1-layer MLP]{\includegraphics[width=0.33\textwidth]{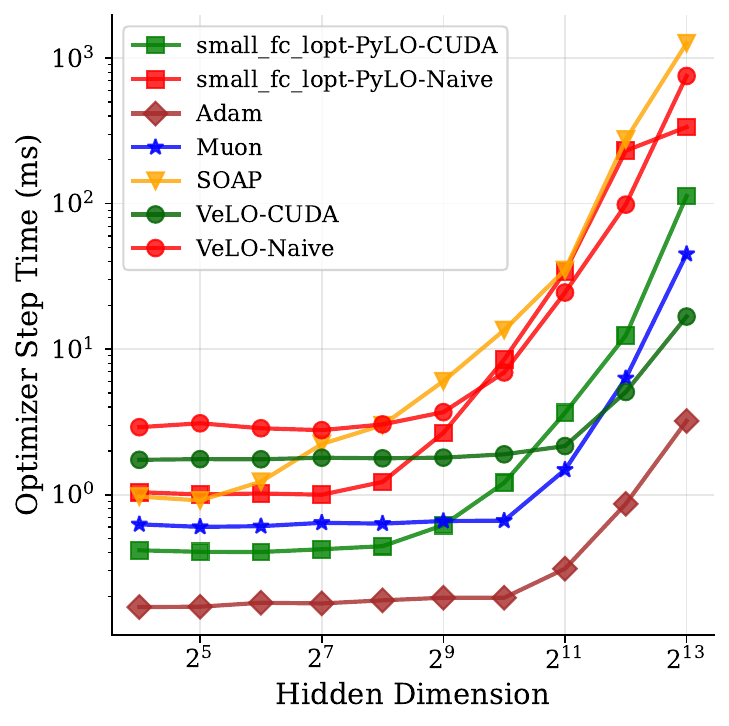}}
    \subfloat[256-width MLP]{\includegraphics[width=0.33\textwidth]{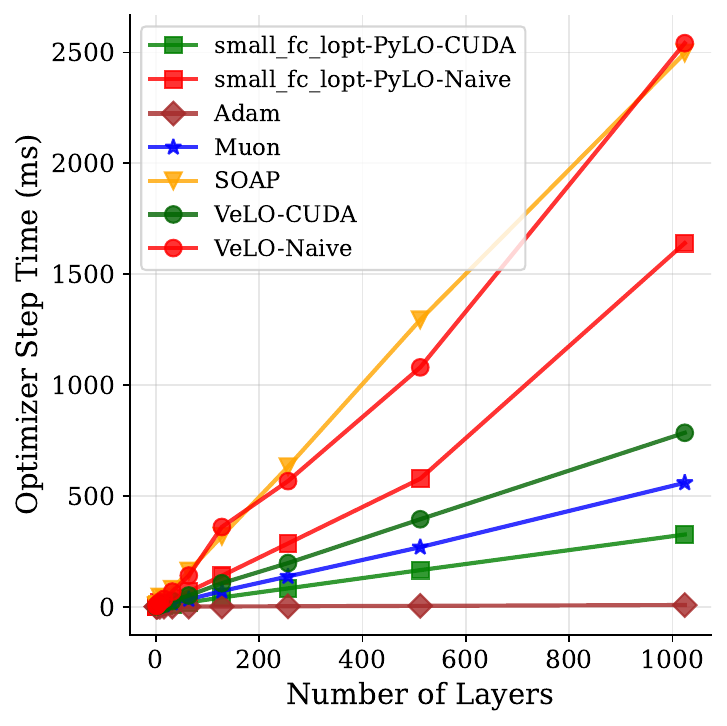}}
    \subfloat[4096-width MLP]{\includegraphics[width=0.33\textwidth]{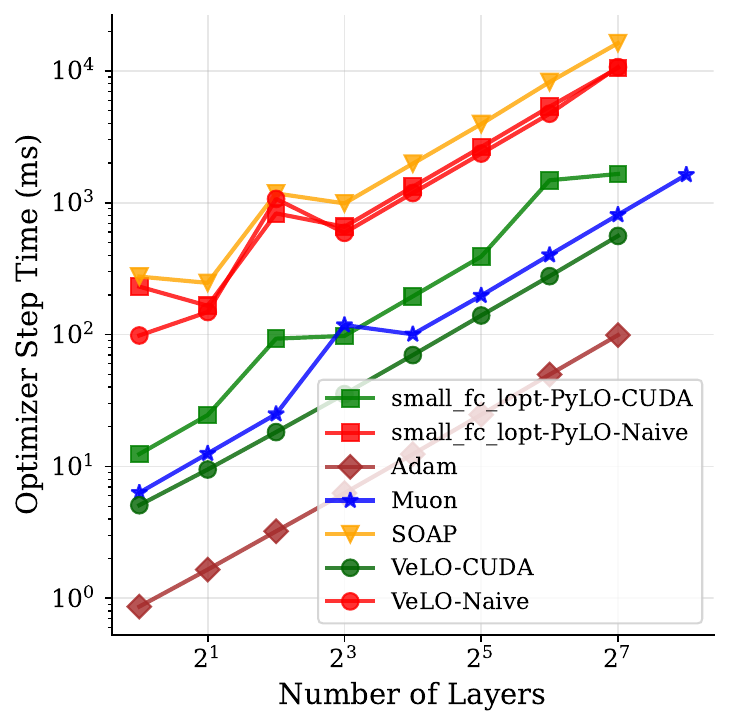}}\\
    \caption{\textbf{Scaling behavior of CUDA implementation with varying MLP configurations}
    We evaluate optimizer step time across various MLPs with controlled width and depth. Subfigure (a) reports optimizer step time as the hidden size of the 1-layer optimizee MLP is increased. Subfigure (b) reports step time as the depth of a 256 hidden-dimension optimizee MLP is increased. Subfigure (c) reports step time as the depth of a $4096$ hidden-dimension optimizee is increased. Missing datapoints correspond to OOM errors. We observe that in each case, our CUDA implementation reduces the optimizer step time by an order of magnitude relative to the naive implementation. 
    }\vspace{-10pt}
    \label{fig:mlp-scaling}
\end{figure*}

\textbf{Scaling behavior of our CUDA implementation vs the JAX baseline} Figure~\ref{fig:scaling-timing} reports the scaling behavior w.r.t.  transformer size when trained with different implementations of \texttt{small\_fc\_lopt} (we use hidden size $32$ as in~\cite{therien2024mulo}) and VeLO\cite{velo}. JAX corresponds to the original implementation of~\cite{metz2022practical}, which uses the JAX compiler, while CUDA is the fastest implementation that we make available in PyLO. We observe that our CUDA implementations results in a substantial memory savings as it can optimize a 1B parameter transformer, while the JAX implementation runs out of memory for an 80 GB A100 GPU. \rev{This is because it materializes the full feature tensor of size $mn \times d_{feat}$ in global memory, which exceeds GPU memory at the 1B scale, whereas the CUDA kernels avoid this intermediate allocation entirely.} Moreover, in cases where the JAX implementation does not run out of memory, it is outperformed by our CUDA implementation, resulting in a $2\times$ reduction in step time.

\textbf{Isolating Scaling behavior of our CUDA implementation with MLP taks} To systematically characterize the scaling behavior of our implementation, we conduct controlled experiments using synthetic MLP architectures with varying numbers of layers and hidden dimensions. This allows us to explicitly control for tensor sizes and the number of tensors to measure the effect on our implementation. Figure~\ref{fig:mlp-scaling} presents three complementary scaling analyses that isolate the effects of network width and depth on optimizer computational overhead.
Figure~\ref{fig:mlp-scaling}(a) examines the relationship between hidden layer dimensionality and optimization time for a fixed 1-layer MLP architecture. The results reveal a stark performance disparity: the naive implementations of learned optimizers exhibit optimization times an order of magnitude greater than our CUDA implementation. While our CUDA implementations remains slower than traditional optimizers, it demonstrates favorable scaling characteristics. We further enhance this on H100 hardware in Appendix \ref{appx:h100}.

Complementary analysis in Figure~\ref{fig:mlp-scaling}(b),(c) evaluates depth scaling behavior using MLPs with fixed width of 256 and 4096 units but varying layer counts (depth). The naive implementation consistently operates in a computationally prohibitive regime, requiring over 400ms for optimization steps, while our CUDA implementation maintains optimization times approaching the performance envelope of traditional optimizers despite the inherent complexity of learned optimization.
These scaling experiments demonstrate that our CUDA implementation successfully addresses the computational bottlenecks that render naive learned optimizer implementations impractical for realistic model architectures.

\section{Distributed Optimizer Step}
\begin{figure}[t]
    \centering
    \includegraphics[width=\linewidth]{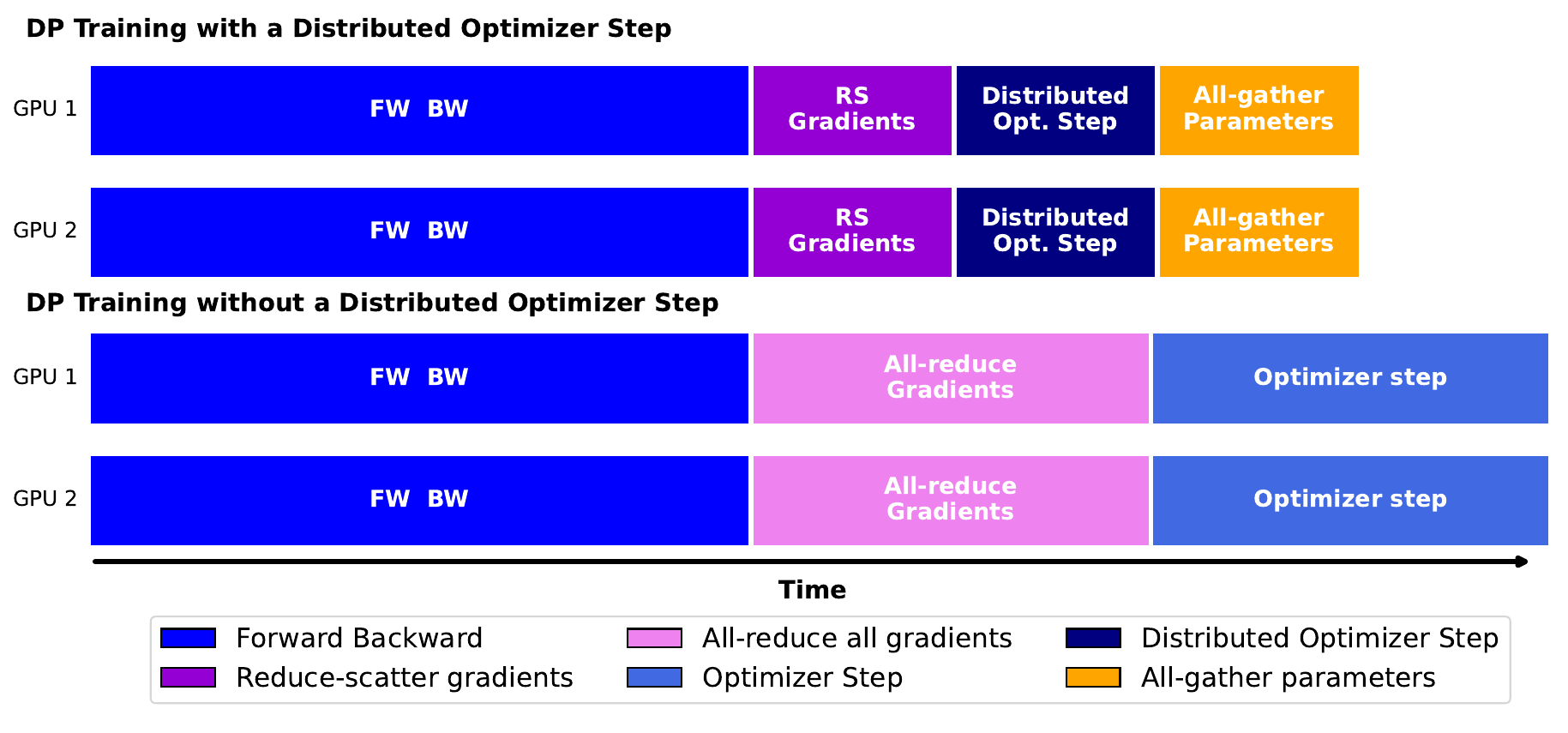}
    \vspace{-24pt}
    \caption{\textbf{Overlapping Optimizer steps with Communication and distributed Optimizer steps.}}
    \label{fig:optcommoverlap}
    \vspace{-15pt}
\end{figure}

It is possible to further reduce optimizer step time by distributing computation across multiple devices. So far, our analysis has focused on optimizer step times for fully replicated data-parallel training where the gradient is first all-reduced and, subsequently, the optimizer step for the entire model is performed \emph{redundantly} and simultaneously on all devices. While not ideal, this approach is reasonable for coordinate-wise optimizers like Adam, which require little computation during the optimizer step. However, optimizers like Shampoo~\cite{gupta2018shampoo}, Muon~\cite{jordan2024muon}, or Learned optimizers~\cite{metz2022practical,velo}, that perform significant computation at each step, can benefit from distributing the optimizer step across multiple devices. As illustrated in figure~\ref{fig:optcommoverlap}, the optimizer step can effectively be distributed across devices by splitting the all-reduce operation into its parts: (1) reduce scatter the gradients evenly for each parameter tensor across devices, (2) perform the optimizer step \emph{non-redundantly} across all devices, and (3) all-gather the updated model parameters.


\begin{table}[t]
\centering
\caption{\textbf{Comparing improvements from distributed optimizer steps.} We report the time taken for the forward and backward pass, communication, and optimizer step for different optimizers when training a 125M transformer language model across four H100 GPUs. All times are reported in milliseconds (ms). The all-reduce column performs an all-reduce of the gradient before the optimizer step, while reduce-scatter distributes the optimizer step across devices (\autoref{fig:optcommoverlap} illustrates the difference between these steps.). The FSDP A2A column uses ZeRO-1 style sharding of optimizer states with all-to-all communication to distribute tensors across devices for the optimizer step. Difference ($\Delta$) columns show improvement relative to the all-reduce baseline. 
}
\label{tab:diststepresults}
\vspace{7pt}
\resizebox{1.0\columnwidth}{!}{%
\begin{tabular}{l|c|cc|cc}
\toprule
\textbf{Optimizer} & All-reduce & Reduce-scatter & $\Delta$ & FSDP A2A & $\Delta$ \\
\midrule
small\_fc\_lopt CUDA & 136.53 & 104.67 & \cellcolor{lightgreen}$-$31.86 & 110.53 & \cellcolor{lightgreen}$-$26.00 \\
VeLO CUDA            & 127.67 & 108.71 & \cellcolor{lightgreen}$-$18.96 & 116.00 & \cellcolor{lightgreen}$-$11.67 \\
Muon                 & 106.92 & 98.10  & \cellcolor{lightgreen}$-$8.82  & 100.92 & \cellcolor{lightgreen}$-$6.00 \\
Adam                 & 99.93  & 95.93  & \cellcolor{lightgreen}$-$4.00  & 96.46  & \cellcolor{lightgreen}$-$3.47 \\
\bottomrule
\end{tabular}
}\vspace{-5pt}
\end{table}

\textbf{Improvements from using a distributed optimizer step} Table~\ref{tab:diststepresults} reports the time taken for the forward and backward pass, communication, and optimizer step for different optimizers when training a 125M transformer language model across four H100 GPUs. We observe that distributing the optimizer step can lead to substantial improvements for all optimizers, but that learned optimizers improve the most from a distributed step: by $31.86$ ms for small\_fc\_lopt and $18.96$ ms for VeLO. This reduces the overhead relative to Adam for these optimizers to just $9\%$ and $13\%$, respectively. These results suggest that, in large batch training settings, the overhead or learned optimizers relative to Adam and Muon can further be reduced by distributing the optimizer step over more devices.

\rev{
\textbf{Sharded Optimizer States.} Our analysis shows that tensor-level reduce-scatter significantly accelerates distributed optimizer steps. Further gains in memory and compute are possible by sharding optimizer states across devices, as in ZeRO~\cite{zero} (i.e., fully sharded data parallelism~\cite{fsdp}). To evaluate this, we implement a ZeRO-1 style optimizer step that shards only optimizer states across devices and uses an all-to-all operation to distribute tensors for the optimizer step. The FSDP A2A column in Table~\ref{tab:diststepresults} reports these results. We find that the FSDP A2A approach provides substantial improvements over the all-reduce baseline for all optimizers, reducing step time by $26.00$ ms for small\_fc\_lopt and $11.67$ ms for VeLO. While the reduce-scatter approach remains faster overall, the FSDP A2A approach offers additional memory savings by sharding optimizer states across devices. Since learned optimizers currently operate per-parameter (except normalization), the costly all-to-all could potentially be replaced with a distributed normalization step that communicates only $d_{\text{feat}}$ scalars. We leave this optimization for future work.}

\begin{figure*}[t]
    \centering
    \subfloat[LM $\mu$LO]{\includegraphics[width=0.25\linewidth]{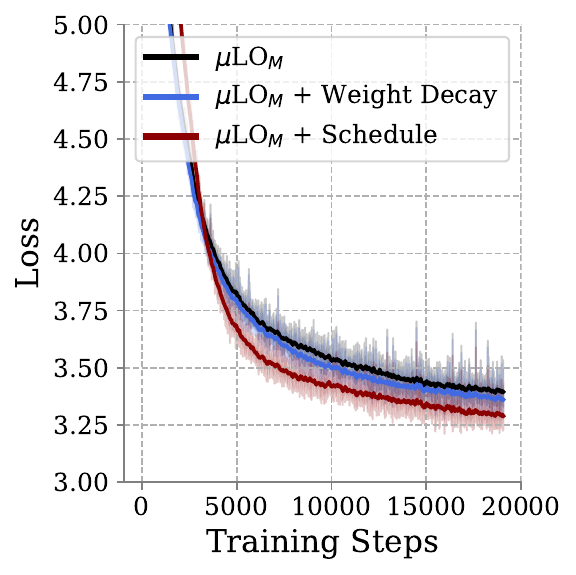}}
    \subfloat[LM VeLO]{\includegraphics[width=0.25\linewidth]{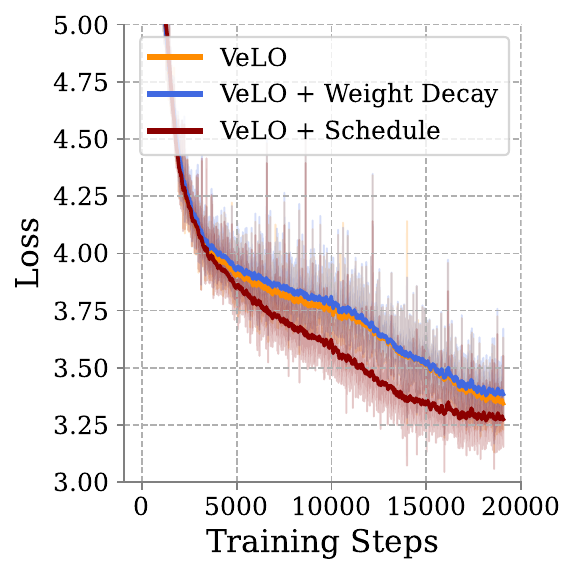}}
    \subfloat[ViT $\mu$LO]{\includegraphics[width=0.25\linewidth]{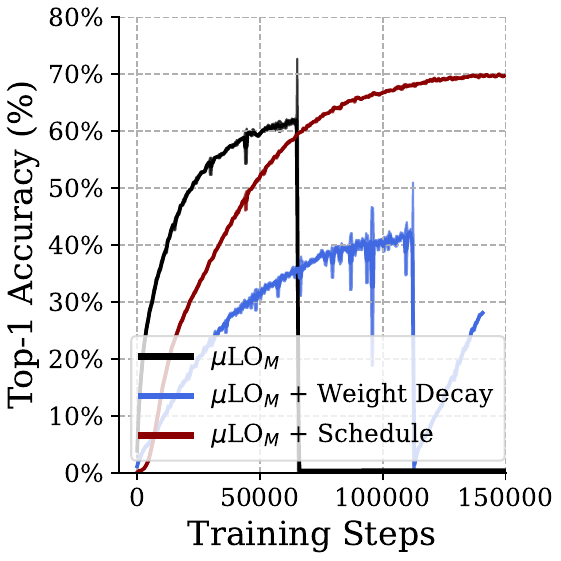}}
    \subfloat[ViT VeLO]{\includegraphics[width=0.25\linewidth]{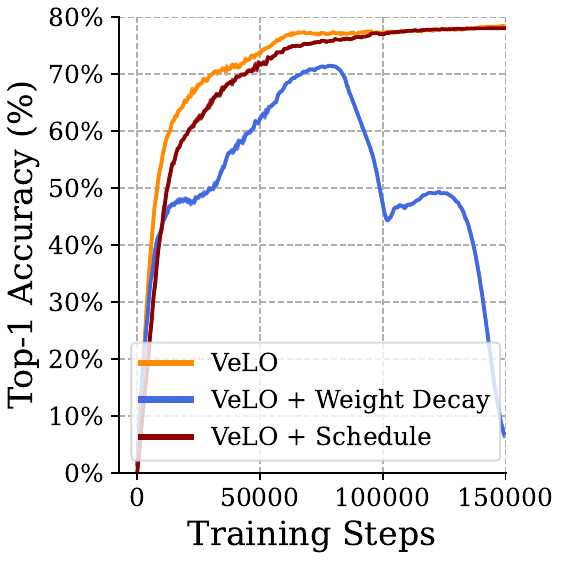}}
    \caption{\textbf{Weight decay and learning rate schedule (LRS) ablation.}  In PyLO, it is easy to equip VeLO and $\mu$LO$_M$ with weight decay and learning rate schedules since it integrates with \texttt{torch.optim.Optimizer}. We sweep different values and find that in some cases learned optimizers improve. 
    }\vspace{-21pt}
    \label{fig:mainfig}
\end{figure*}

\section{Demonstrating the real-world use of PyLO}
In the following section, we evaluate and compare our PyTorch implementation of VeLO~\cite{velo} and \texttt{small\_fc\_lopt} to the performance of tuned Adam with a cosine annealing schedule. We use $\mu$LO$_M$~\cite{therien2024mulo} pretrained weights for the \texttt{small\_fc\_lopt}. Our goal is to establish the performance of readily available learned optimizers in PyLO for these practical vision and language pre-training tasks.

\subsection{Training Vision Transformer on ImageNet-1K}
\textbf{Experimental Configuration:} We train ViT-Base/16 models (86M parameters) for 480 epochs (150k steps) on ImageNet-1K following established protocols \cite{rw2019timm}. We apply standard augmentation techniques (RandAugment, CutMix, Mixup) to ensure realistic training conditions that reflect practical deployment scenarios. We employ a batch size of 4096 and conduct these experiments on Nvidia H100 GPUs.

\textbf{Results} As shown in \autoref{tab:final_loss_medium_only} VeLO demonstrates competitive performance in this context, achieving 78.45\% top-1 accuracy compared to Adam's 77.22\%. This scenario demonstrates a case where the learned optimizer practically competes with Adam. $\mu$LO exhibits training divergence after 65,000 steps due to its limited 1,000-step meta-training horizon, achieving only 62.14\% peak validation accuracy. 

\begin{table}[h]
\centering
\caption{Final Loss for Language Modeling (LM) and Final Validation Accuracy for Vision Transformer (ViT) Tasks for Medium Model Sizes Using Different Optimizers}
\label{tab:final_loss_medium_only}
\resizebox{1.0\columnwidth}{!}{%
\begin{tabular}{l|c|c}
\toprule
\textbf{Optimizer} & \textbf{LM Loss $\downarrow\;$(355M) } & \textbf{ViT Acc. $\uparrow\;$(86M)} \\
\midrule
$\mu$LO$_M$         & 3.18 & 62.14 \\
VeLO                & \textbf{2.89} & \textbf{78.39} \\
Adam + Cosine      & 2.91 & 77.22 \\
\bottomrule
\end{tabular}
}
\end{table}

\subsection{Language model pre-training on FineWeb}
\textbf{Experimental Configuration}\\
We pre-train a $355$M parameter GPT-2 style transformers on FineWeb-EDU~\cite{lozhkov2024fineweb-edu} for $10$B tokens, employing a batch size of $512$ and sequences of length $1024$ (see section~\ref{apdx:sec:hparams} for additional hyperparameter details).r more details.  \\

\textbf{Results}\\
Table~\ref{tab:final_loss_medium_only} reports the final loss achieved for all optimizers in our study, while figure~\ref{apdx:fig:gpt} of the appendix plots the training curves. We observe that VeLO outperforms Adam and $\mu$LO$_M$, reaching the lowest final loss. This demonstrates that PyLO has the potential to deliver optimization performance improvements to PyTorch users.

\subsection{Combining scheduler and decoupled weight decay}
To showcase the interoperability of our library, we add learning rate scheduling and decoupled weight decay to our learned optimizer implementations in as little as $5$ lines of code. Although the primary objective of learned optimization is to eliminate the reliance on manually designed heuristics, we take the opportunity to investigate whether learning rate schedules and weight decay can still improve learned optimizer performance.

\textbf{Effect of Learning Rate Scheduler}\\
We equip $\mu$LO$_M$ and VeLO with a cosine annealing LRS and tune the maximum learning rate (MaxLR). Figure~\ref{fig:mainfig} reports ViT and GPT training and accuracy curves for the best-performing MaxLR while training curves for more values are reported in the appendix (\ref{appen:extended-ablations}). We observe that $\mu$LO$_M$ benefits substantially from explicit scheduling, extending stable training from 65k to 150k steps and reaching 71\% Top-1 accuracy on ImageNet and improving loss in language model pre-training. VeLO shows minimal scheduling improvements, suggesting different internal adaptation mechanisms.

\textbf{Effect of Weight decay} \\
We equip $\mu$LO$_M$ and VeLO with decoupled weight decay and tune the decay coefficient, $\lambda$. Figure~\ref{fig:mainfig} reports ViT and GPT training and accuracy curves for the best-performing $\lambda$ while training curves for more values are reported in the appendix (\ref{appen:extended-ablations}). We observe that $\mu$LO$_M$ benefits from weight decay for training GPT but not VeLO. Neither optimizer benefits from weight decay for ViT tasks.

\section{Conclusion}
We have presented PyLO, an open-source PyTorch library that removes key barriers to the practical use of learned optimizers in large-scale deep learning. PyLO offers drop-in implementations of state-of-the-art methods such as VeLO and \texttt{small\_fc\_lopt} through the standard \texttt{torch.optim.Optimizer} interface, with seamless Hugging Face Hub integration for model sharing. Central to our systems contributions are custom CUDA kernels that fuse feature construction, normalization, and MLP inference, and a distributed implementation that distributes tensors across multiple devices. This achieves 86--88\% lower overhead than naive PyTorch, and can efficiently scale to training a 1B-parameter transformer on a single A100 GPU. Our evaluation shows that PyLO reduces learned optimizer overhead to acceptable levels at scale and they can deliver competitive accuracy on ImageNet and large-scale language pretraining. As future work, we envision PyLO enabling  (1) hardware-optimizer co-design (2) behavior-cloned distillation of expensive second-order optimizers such as SOAP\cite{vyas2025soap}. PyLO thus establishes a robust foundation for community-driven research and practical deployment of learned optimizers in modern ML systems.

\clearpage

\bibliography{mlsys/ref}
\bibliographystyle{mlsys/mlsys2025}

\clearpage

\appendix
\onecolumn

\section{Artifact Appendix}

\subsection{Abstract}

PyLO is a learned optimizer library that provides an accessible way to load and run pretrained learned optimizers in downstream tasks.
It follows the standard \texttt{torch.optim.Optimizer} interface, enabling drop-in replacement of conventional optimizers such as Adam or AdamW with minimal changes to existing training code.
PyLO also provides fast custom CUDA kernels that substantially reduce optimizer step time compared to naive PyTorch or JAX implementations.
The core library is available at \url{https://github.com/Belilovsky-Lab/pylo}, and a companion repository of standard ML workflow examples is available at \url{https://github.com/Belilovsky-Lab/pylo_examples}.
The latter provides reproducible training scripts for image classification (ViT-B/16 on ImageNet) and language model pre-training (GPT-2-style transformers), enabling direct comparison of learned optimizers against baselines such as AdamW with cosine annealing.

\subsection{Artifact Checklist}

\begin{itemize}
    \item \textbf{Algorithm:} Learned optimizer inference (\texttt{small\_fc\_lopt}, VeLO, \textmu LO) with CUDA-accelerated optimizer steps.
    \item \textbf{Program:} Python library (\texttt{pylo}) and example training scripts (\texttt{pylo\_examples}).
    \item \textbf{Compilation:} CUDA toolkit required for custom kernel build; \texttt{CUDA\_HOME} must be set. CUDA version of PyTorch must match \texttt{nvcc} version.
    \item \textbf{Binary:} No precompiled binaries are distributed; the package is built from source.
    \item \textbf{Data set:} ImageNet (for ViT experiments); standard language modeling datasets (for GPT experiments). Datasets are not bundled; users must supply them.
    \item \textbf{Run-time environment:} Linux, Python 3.11, PyTorch (CUDA-enabled), \texttt{huggingface\_hub}. Optional: Weights \& Biases for logging.
    \item \textbf{Hardware:} NVIDIA GPU (experiments in the paper use an A100 80\,GB). CPU-only installation is possible but CUDA kernels will not be available.
    \item \textbf{Execution:} Single- or multi-GPU training via \texttt{torchrun}.
    \item \textbf{Metrics:} Optimizer step time (ms), training throughput (samples/sec), validation loss / accuracy.
    \item \textbf{Output:} Training logs (optionally synced to Weights \& Biases), model checkpoints.
    \item \textbf{Experiments:} Step-time comparisons; image classification and language model pre-training comparisons.
    \item \textbf{How much disk space required (approximately)?} $\sim$5\,GB (excluding datasets).
    \item \textbf{How much time is needed to prepare the workflow (approximately)?} 15--30 minutes (installation and environment setup).
    \item \textbf{How much time is needed to complete experiments (approximately)?} Step-time benchmarks: $<$1 hour. Full ViT or GPT training runs: several hours to days depending on hardware.
    \item \textbf{Publicly available?} Yes.
    \item \textbf{Code licenses:} BSD-3-Clause (\texttt{pylo}); Apache-2.0 (\texttt{pylo\_examples}).
\end{itemize}

\subsection{Description}

\subsubsection{How to Access}

The PyLO library is publicly available at:
\begin{center}
    \url{https://github.com/Belilovsky-Lab/pylo}
\end{center}
The companion examples repository (training scripts for ViT and GPT workflows) is available at:
\begin{center}
    \url{https://github.com/Belilovsky-Lab/pylo_examples}
\end{center}
Documentation is hosted at \url{https://belilovsky-lab.github.io/pylo}. Pre-trained Learned optimizer weights are distributed via Hugging Face Hub and are downloaded automatically on first use (default \texttt{hf-keys} provided).

\subsubsection{Hardware Dependencies}

An NVIDIA GPU is required to build and use the custom CUDA kernels. The benchmarks reported in the paper were performed on a single NVIDIA A100 80\,GB GPU. The library can be installed without CUDA kernels (CPU-only or standard PyTorch GPU), but optimizer step times will be significantly slower.

\subsubsection{Software Dependencies}

\begin{itemize}
    \item Python 3.11
    \item PyTorch (CUDA-enabled build; CUDA version must match the installed \texttt{nvcc})
    \item \texttt{huggingface\_hub} (for automatic weight download)
    \item CUDA Toolkit (\texttt{CUDA\_HOME} must be set for kernel compilation)
    \item \textit{Optional:} Weights \& Biases (\texttt{wandb}) for experiment logging
    \item \textit{Optional:} Custom \texttt{mup} fork for \textmu P / \textmu LO support: \url{https://github.com/bentherien/mup}
\end{itemize}

\subsubsection{Data Sets}

\begin{itemize}
    \item \textbf{ImageNet (ILSVRC-2012):} Required for ViT-B/16 image-classification experiments. Must be obtained independently and the path supplied via \texttt{--data-dir}.
    \item \textbf{Language modeling datasets:} Required for GPT pre-training experiments. Configuration is specified inside \texttt{./language\_model\_pretraining}.
\end{itemize}

\subsection{Installation and Evaluation}

\subsubsection{Installation}

\paragraph{Quick install (no CUDA kernels).}
This variant uses standard PyTorch operations and requires no compilation step:
\begin{verbatim}
pip install git+https://github.com/Belilovsky-Lab/pylo
\end{verbatim}

\paragraph{Build from source with custom CUDA kernels (recommended).}
CUDA must be installed and \texttt{CUDA\_HOME} must point to the CUDA installation. The CUDA version of PyTorch must match \texttt{nvcc}:
\begin{verbatim}
git clone https://github.com/Belilovsky-Lab/pylo
cd pylo
pip install .
python setup.py install --cuda
\end{verbatim}
Alternatively:
\begin{verbatim}
pip install --no-build-isolation \
    --config-settings="--build-option=--cuda" .
\end{verbatim}

\paragraph{Optional: MuP patch for \textmu LO support.}
After installing the library, apply the MuP patch:
\begin{verbatim}
python -m pylo.util.patch_mup
\end{verbatim}

\paragraph{Setting up the examples repository.}
A self-contained setup using Miniconda is provided:
\begin{verbatim}
install_dir=$PWD/pylo_install
mkdir $install_dir && cd $install_dir

wget https://repo.anaconda.com/miniconda/\
Miniconda3-py311_24.7.1-0-Linux-x86_64.sh
bash Miniconda3-py311_24.7.1-0-Linux-x86_64.sh \
    -b -p $PWD/miniconda3
source $PWD/miniconda3/bin/activate

pip install git+https://github.com/bentherien/mup.git

git clone https://github.com/Belilovsky-Lab/pylo.git
cd pylo && pip install .
python setup.py install --cuda
\end{verbatim}
For Weights \& Biases logging, set the following environment variables:
\begin{verbatim}
export WANDB_API_KEY=YOUR_KEY
export WANDB_PROJECT=pylo_examples
export WANDB_MODE=online
\end{verbatim}

\subsubsection{Basic Usage}

PyLO optimizers conform to the \texttt{torch.optim.Optimizer} interface. The only difference from standard PyTorch optimizers is that for some optimizers the current loss value must be passed to \texttt{optimizer.step()}:
\begin{verbatim}
import torch
from pylo.optim import VeLO_CUDA

model = torch.nn.Linear(10, 2)
optimizer = VeLO_CUDA(model.parameters())

for epoch in range(num_epochs):
    optimizer.zero_grad()
    loss = loss_fn(model(inputs), targets)
    loss.backward()
    optimizer.step(loss)  # loss value required for VeLO
\end{verbatim}

\clearpage
\section{Enhanced Performance on NVIDIA H100 GPUs}
\label{appx:h100}
To further demonstrate the scalability of our approach on state-of-the-art hardware, we evaluate its performance on NVIDIA H100 GPUs, which offer advanced features. These accelerators are particularly beneficial for large matrix operations, such as those with dimensions $8192 \times 8192$---corresponding to the hidden size of a 70B-parameter transformer model.

In our enhanced implementation (denoted \texttt{small\_fc\_lopt-PyLO-CUDA++}), we introduce the following optimizations:
\begin{itemize}
    \item Reduction in the number of global MLP reads to minimize HBM accesses;
    \item Asynchronous memory copies to overlap data movement with computation; and
    \item Full utilization of tensor cores and WGMMA instructions for high-throughput matrix multiplications on large tensors.
\end{itemize}

\begin{figure}[H]
    \centering
    \includegraphics[width=0.85\linewidth]{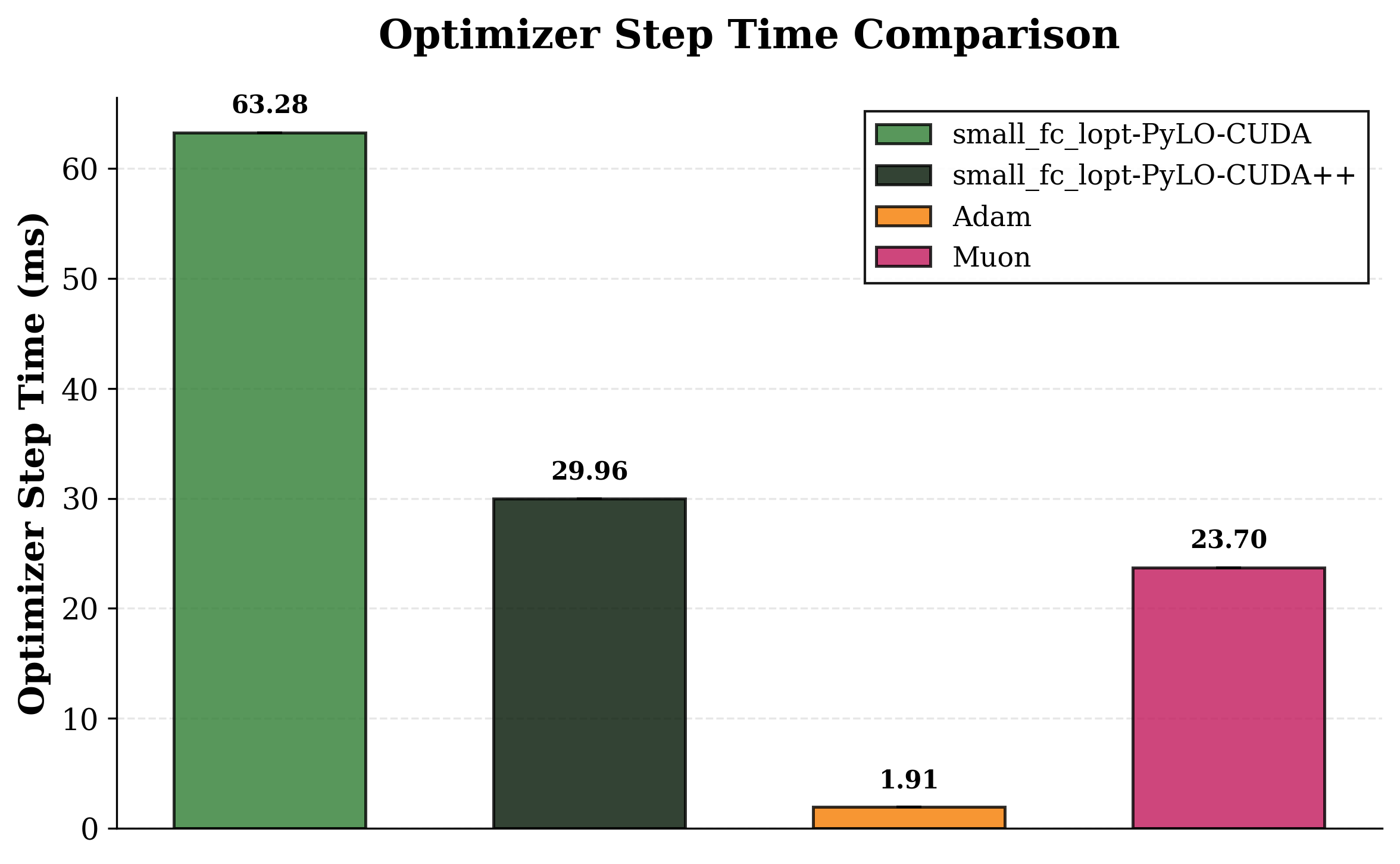}
    \caption{Optimizer step time comparison on NVIDIA H100 GPUs for $8192 \times 8192$ matrices. Our enhanced PyLO-CUDA++ implementation (dark green) achieves a $2.1\times$ speedup over the baseline PyLO-CUDA (light green). Muon (pink) and Adam (orange) given for reference}
    \label{fig:h100_improvements}
\end{figure}

These modifications substantially reduce the optimizer step time, effectively closing the performance gap relative to state-of-the-art baselines for large-scale workloads. The results are summarized in Figure~\ref{fig:h100_improvements}.

\section{Training Curves for the Experiments}
\label{appen:extended-vit-experiments}
\subsection{Performance v.s. Steps}
\begin{figure*}[ht]
    \centering
     \subfloat[ViT small patch 16]{\includegraphics[width=0.5\linewidth]{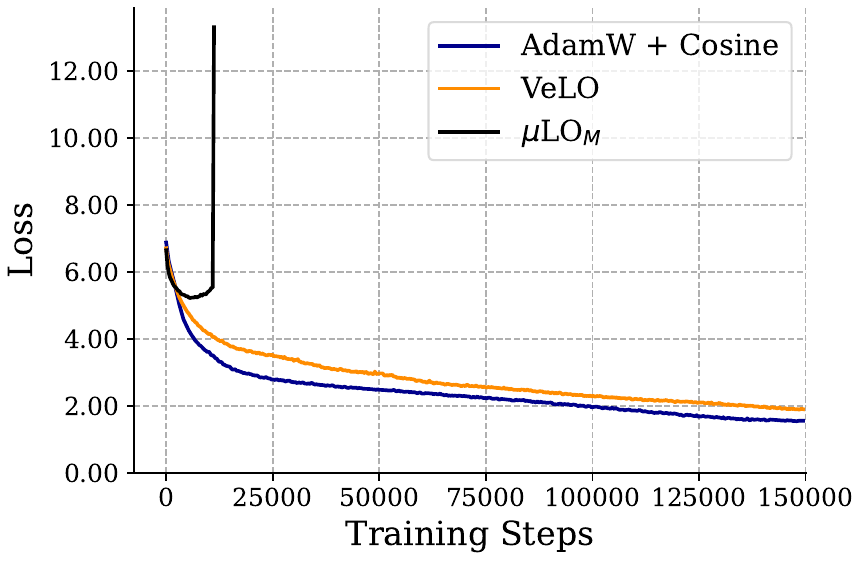}}
    \subfloat[ViT base patch 16]{\includegraphics[width=0.5\linewidth]{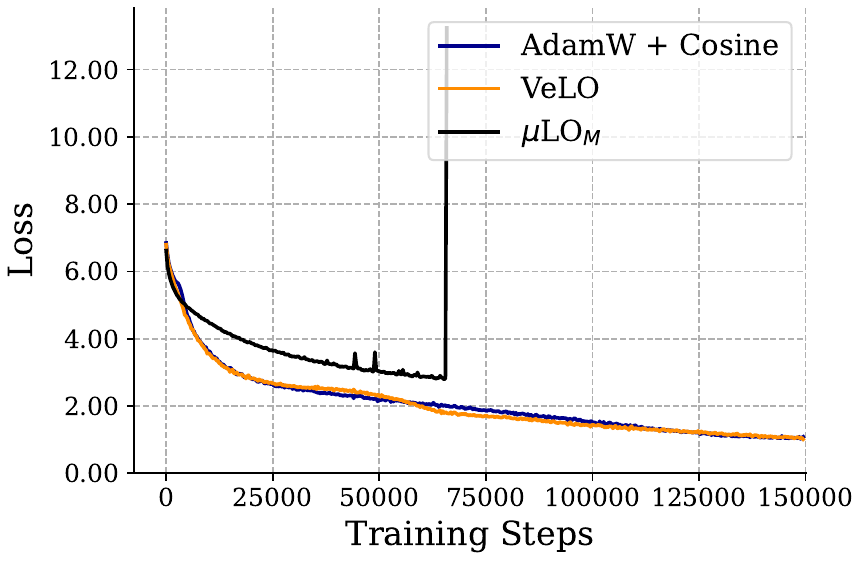}}\\
     \subfloat[ViT large patch 16]{\includegraphics[width=0.5\linewidth]{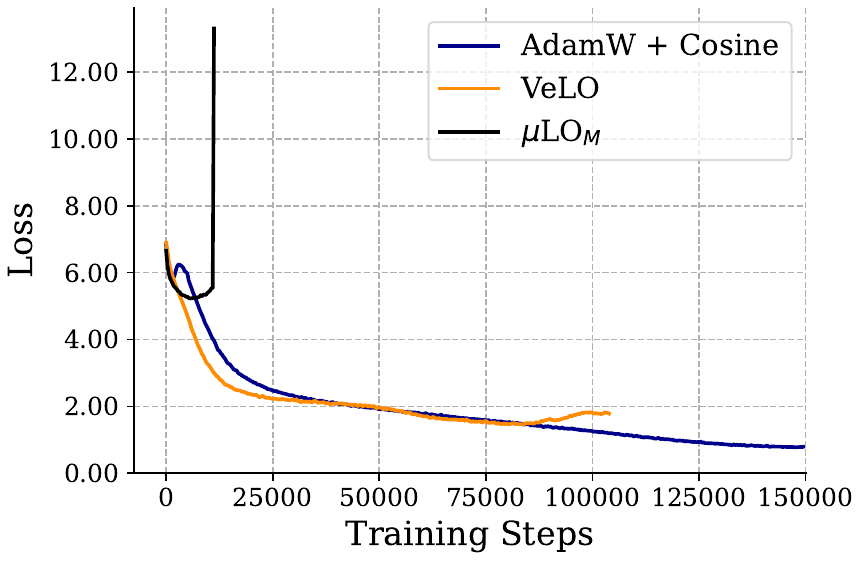}}
    \caption{\textbf{Vision Transformer training dynamics.} Training loss curves for Vision Transformers trained on ImageNet-1k classification across three model configurations: (a) ViT-small with patch size 16, (b) ViT-base with patch size 16, and (c) ViT-large with patch size 16. We compare Adam with cosine learning rate scheduling (blue), VeLO learned optimizer (orange), and $\mu$LO learned optimizer (black) over 150,000 training steps with batch size 4,096. $\mu$LO exhibits early training instability across all scales, while VeLO demonstrates competitive or superior convergence properties, particularly evident in the base model configuration where it achieves lower final loss than the Adam baseline.}
    \label{appen:fig:vit-extended}
\end{figure*}

\begin{figure*}[h]
    \centering
    \subfloat[355M LM]{\includegraphics[width=0.5\linewidth]{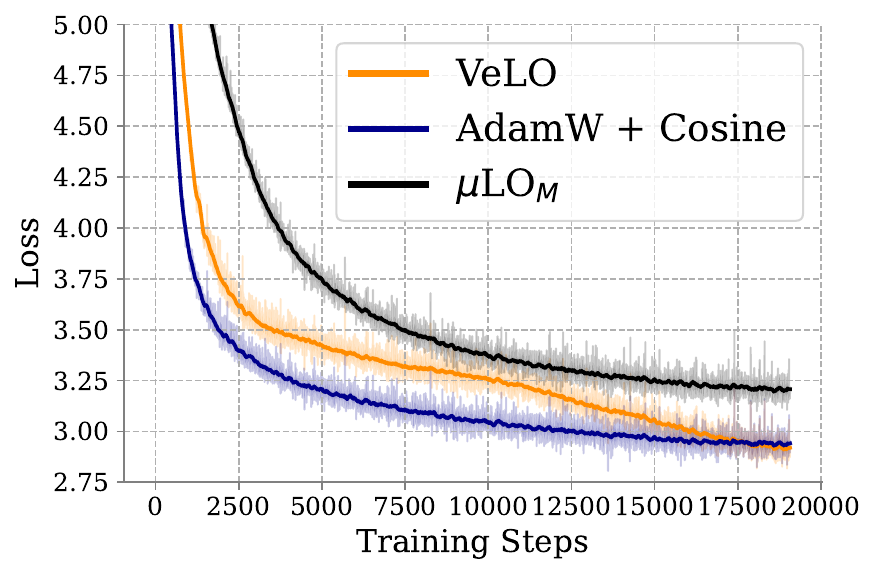}}
    \caption{\textbf{GPT-2 pretraining performance.} We pre-train decoder-only transformers of different sizes on a causal language modeling objective. We estimate gradients from 512 sequences of length 1024, resulting in a $\sim$0.5M token batch size per step. We train all models for 10B tokens of FineWeb-EDU data.}
    \label{apdx:fig:gpt}
\end{figure*}

This section presents a comprehensive empirical evaluation comparing learned optimizers VeLO \cite{velo} and $\mu$LO \cite{therien2024mulo} against the standard Adam optimizer \cite{adamw,kingma2017adam} with cosine learning rate scheduling \cite{warmrestarts}. Our evaluation spans Vision Transformer architectures of varying scales and GPT-2 models, extending beyond the main paper's validation accuracy results for ViT-Base/16 and final loss metrics for GPT-2 to provide detailed training dynamics across multiple configurations.\\

\textbf{Vision Transformer Experiments:} We evaluate three ViT configurations—Small, Base, and Large with patch size 16—on ImageNet-1k classification tasks. All configurations maintain consistent training protocols with 150,000 optimization steps and batch sizes of 4,096 across model variants, using hyperparameters detailed in Table \ref{table:vit-hyperparameters}.
The extended evaluation presented in Figure \ref{appen:fig:vit-extended} reveals critical scale-dependent optimization characteristics. MuLO demonstrates pronounced early divergence patterns across both small and large model configurations, reinforcing our hypothesis that its constrained meta-training horizon limits generalization to extended optimization trajectories. In contrast, VeLO exhibits remarkable consistency in convergence properties across the full spectrum of model scales examined, maintaining stable training dynamics while achieving competitive performance metrics relative to the Adam baseline.\\

\textbf{Language Modeling Experiments}
For GPT-2 pretraining on FineWeb-EDU data, we train models for 19,073 steps with batch size 512 and sequence length 1024. Complete hyperparameters are presented in Table \ref{appen:table:gpt2-hyperparameters}, with full training curves shown in Figure \ref{apdx:fig:gpt}. VeLO demonstrates competitive performance against Adam with cosine scheduling, while MuLO exhibits significantly degraded performance across the extended training horizon, consistent with the stability patterns observed in vision tasks.
\subsection{Performance v.s. Wall-clock time}

\begin{figure*}[ht]
    \centering
     \includegraphics[width=0.5\linewidth]{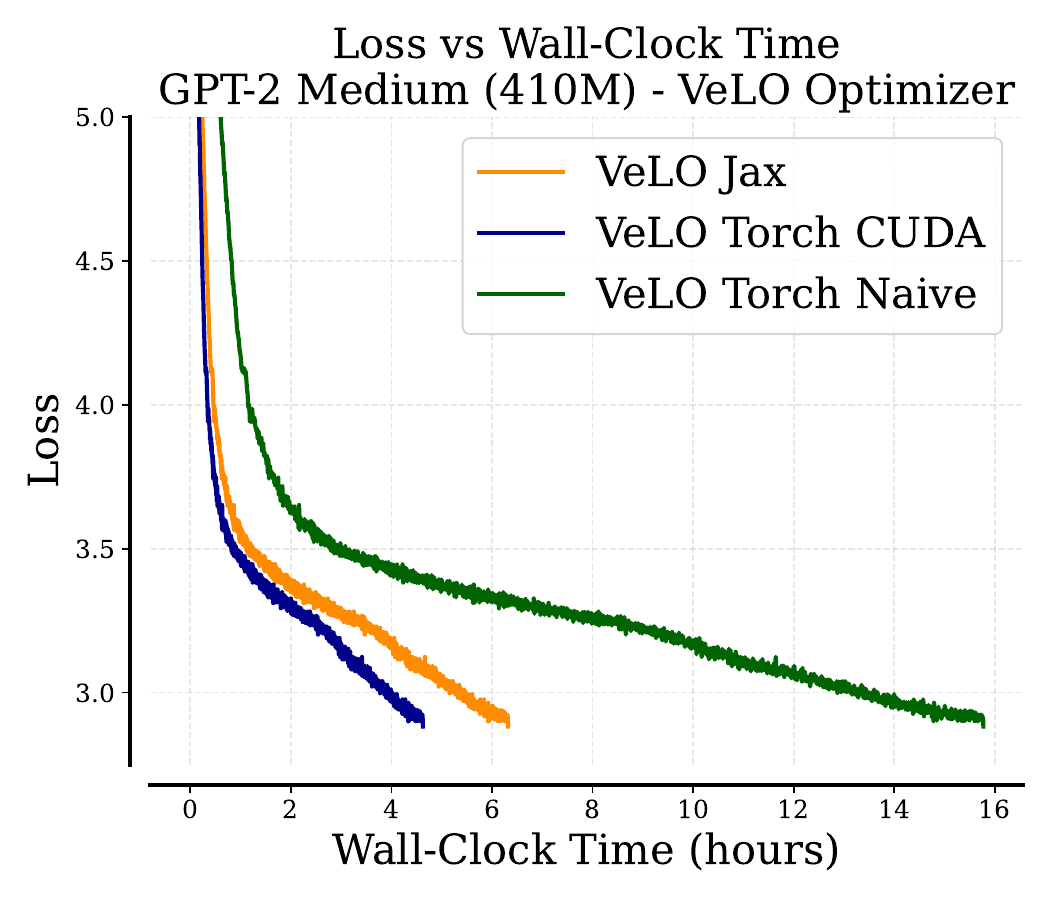}
    \caption{\textbf{GPT2 Medium wall-clock training time.}}
    \label{appen:fig:vit-extended}
\end{figure*}

\begin{figure*}
    \centering
     \includegraphics[width=0.5\linewidth]{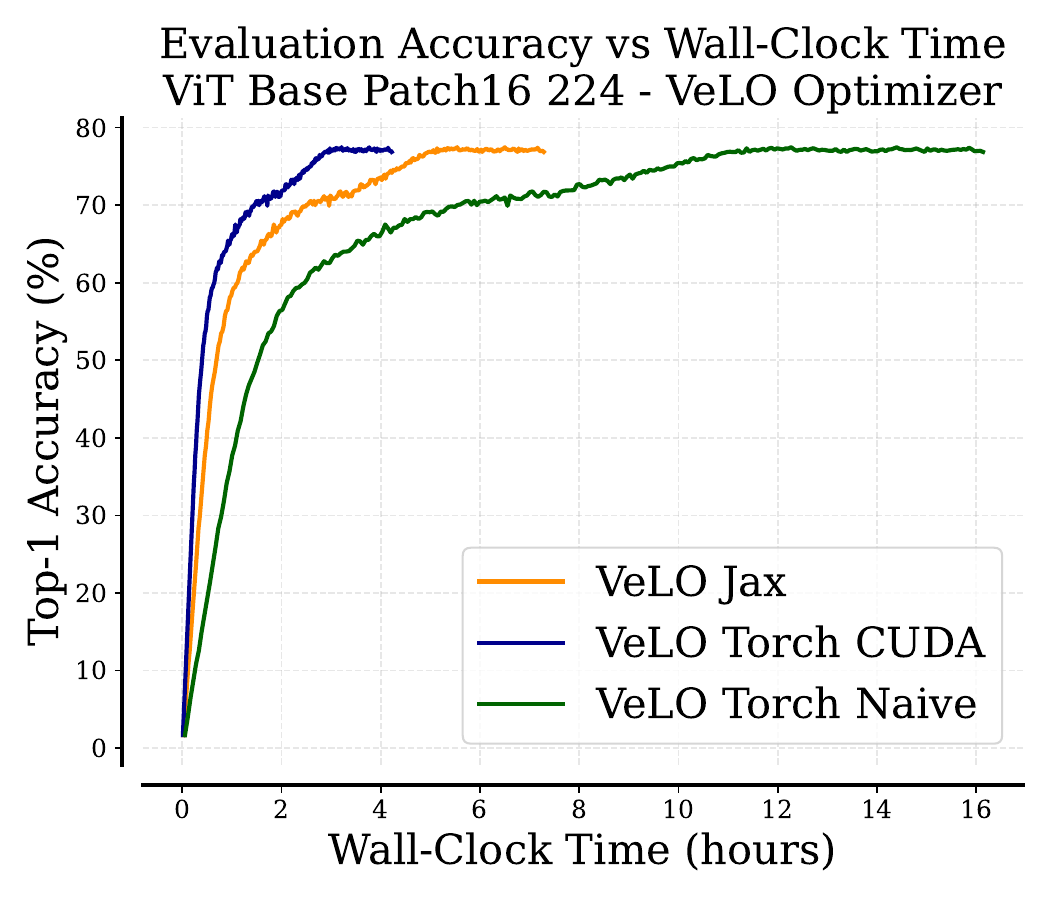}
    \caption{\textbf{ViT Base patch 16 wall-clock training time.}}
    \label{appen:fig:vit-wall_clock}
\end{figure*}

\clearpage

\section{Extended Ablation Studies}
\label{appen:extended-ablations}
This section presents comprehensive training curves across all hyperparameter configurations examined in our weight decay and cosine annealing ablations. Our analysis encompasses systematic sweeps of weight decay coefficients and maximum learning rates for cosine annealing schedules. \\

\begin{figure*}[h]
    \centering
    \subfloat[VeLO Decoupled Weight decay ablation]{\includegraphics[width=0.7\linewidth]{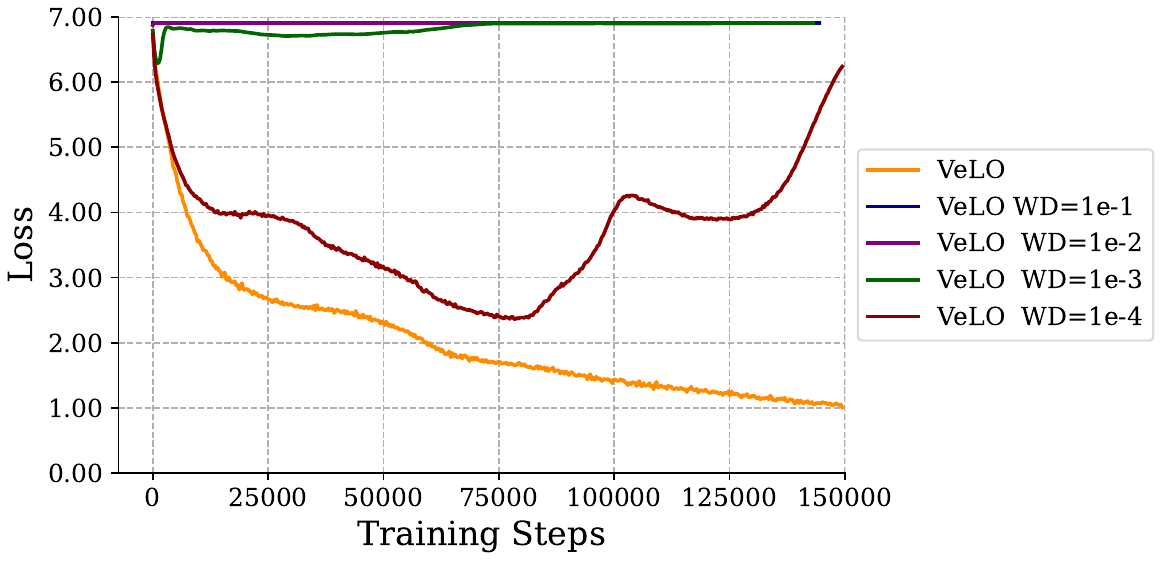}}\\
    \subfloat[VeLO Cosine annealing  Ablation]{\includegraphics[width=0.7\linewidth]{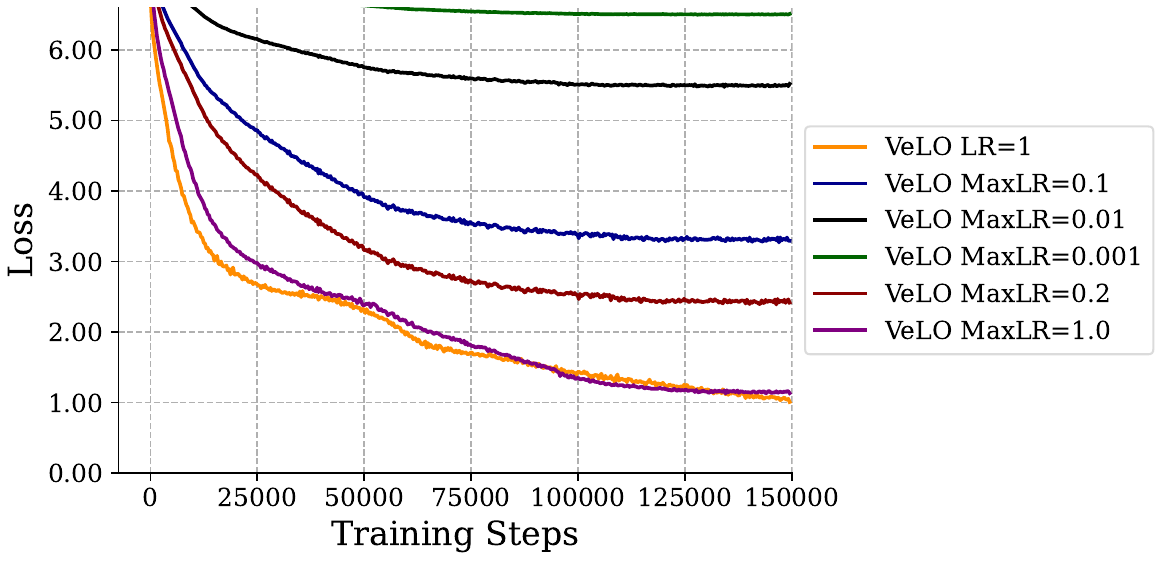}}\\
    \caption{\textbf{Training ViT-B-16  with VeLO on Imagenet 1k with decoupled weight decay and cosine annealing.} We see no improvement in using weight decay or Scheduler for VeLO in this setup}
    \label{appen:fig:velo-extended-wd-lr}
\end{figure*}

\begin{figure*}[h]
    \centering
    \subfloat[$\mu$LO Decoupled Weight decay ablation]{\includegraphics[width=0.7\linewidth]{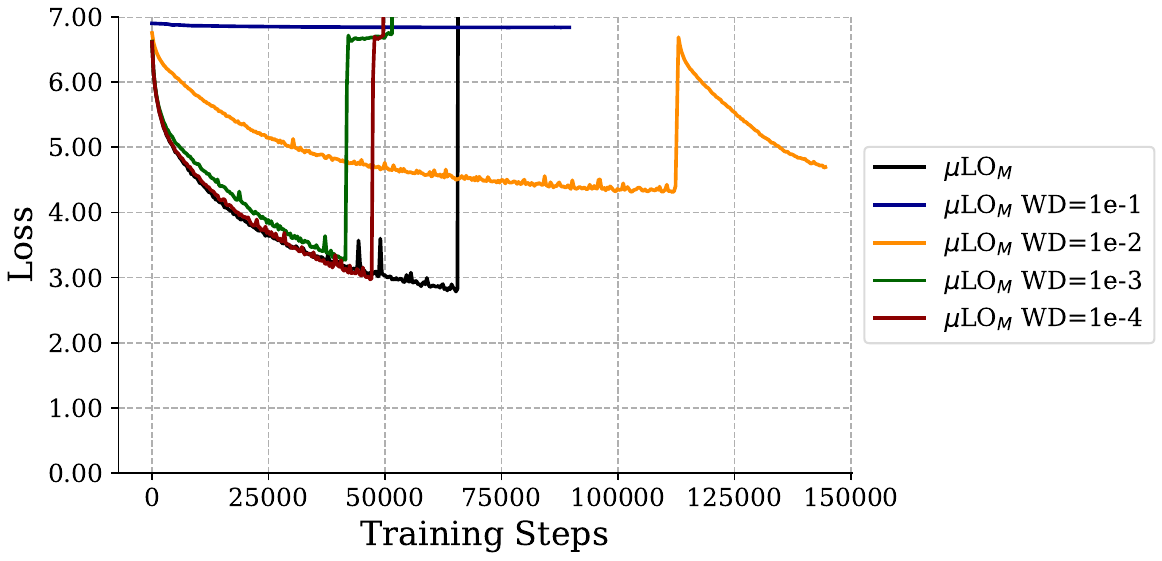}}\\
    \subfloat[$\mu$LO Cosine annealing  Ablation]{\includegraphics[width=0.7\linewidth]{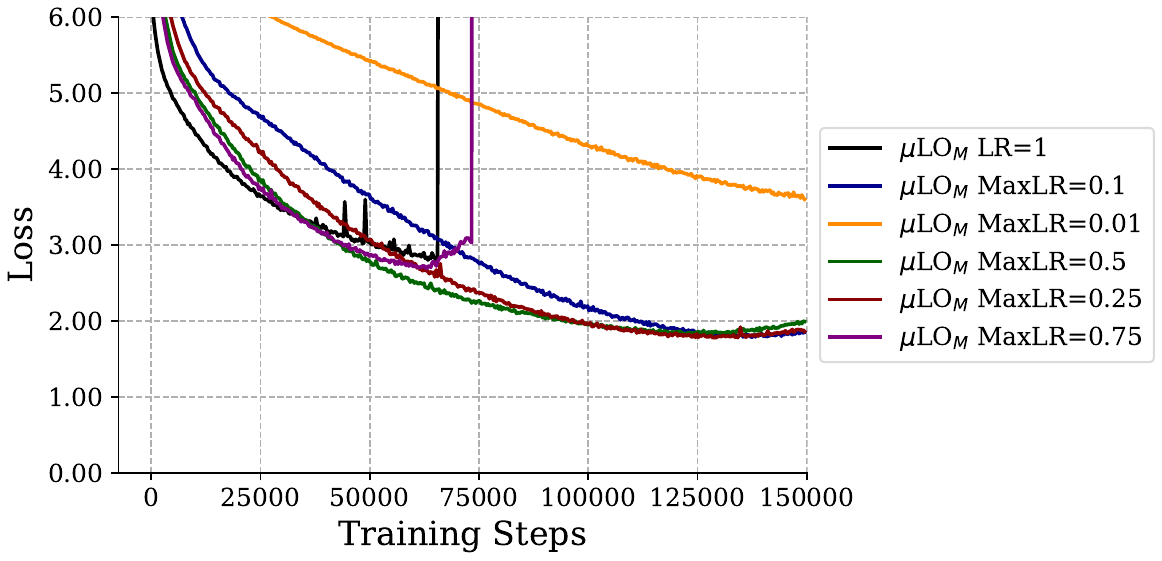}}\\
    \caption{\textbf{ Training ViT-B/16 with $\mu$LO on Imagenet1k with decoupled weight decay and cosine annealing.} We see that the training horizon of $\mu$LO is improved with using a scheduler that helps it to run up to 150000 training steps. We also see that weight decay did not have substantial improvement}
    \label{appen:fig:mulo-extended-wd-lr}
\end{figure*}

\textbf{Vision Transformer Ablations}
Figures \ref{appen:fig:velo-extended-wd-lr} and \ref{appen:fig:mulo-extended-wd-lr} demonstrate the variation of training dynamics with varying maximum learning rates and weight decay values for the ViT-B/16 model. The results reveal that VeLO exhibits minimal sensitivity to weight decay and scheduler modifications in this experimental setup, suggesting robust inherent regularization properties. Conversely, MuLO shows improved training horizon stability when augmented with cosine scheduling, enabling successful completion of the full 150,000 training steps. Weight decay modifications yield marginal improvements for MuLO across the evaluated parameter ranges.\\

\textbf{Language Modeling Ablations}

\begin{figure*}[h]
    \centering
    \subfloat[VeLO Decoupled Weight decay ablation]{\includegraphics[width=0.7\linewidth]{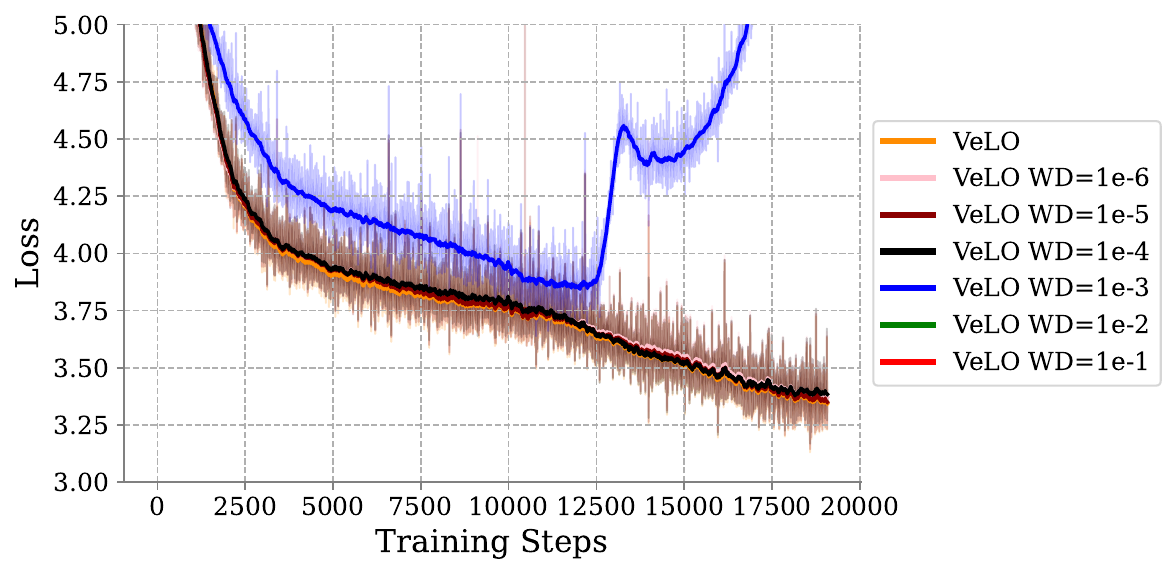}}\\
    \subfloat[VeLO Cosine annealing  Ablation]{\includegraphics[width=0.7\linewidth]{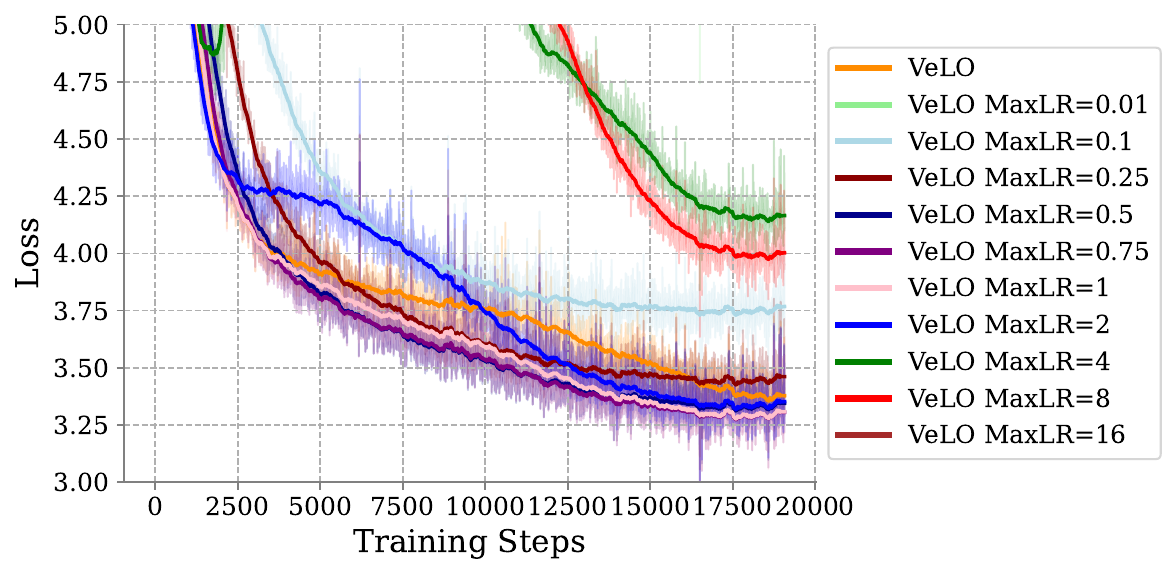}}\\
    \caption{\textbf{GPT2 Pre-training VeLO ablations: decoupled weight decay and cosine annealing.} We pre-train decoder-only transformers of different sizes on a causal language modeling objective. We estimate gradients from $512$ sequences of length $1024$, resulting in a $\sim 0.5$ M token batch size per step. We train all models for $10B$ tokens of FineWeb-EDU data. }
    \label{apdx:fig:gpt-velo-wd-lr}
\end{figure*}

\begin{figure*}[h]
    \centering
    \subfloat[$\mu$LO Decoupled Weight decay ablation]{\includegraphics[width=0.7\linewidth]{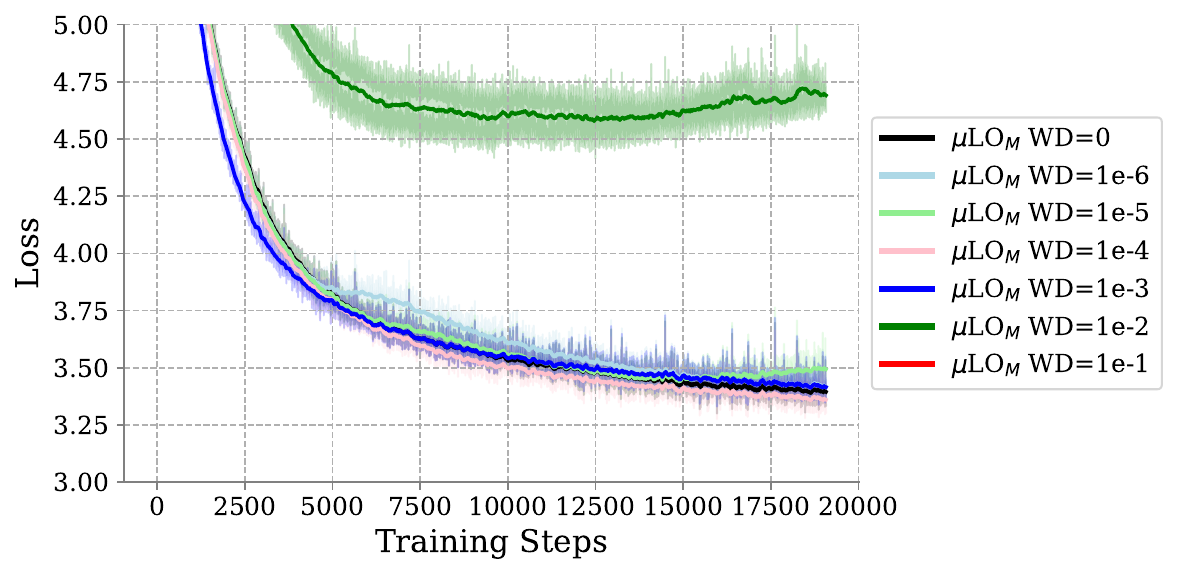}}\\
    \subfloat[$\mu$LO Cosine annealing  Ablation]{\includegraphics[width=0.7\linewidth]{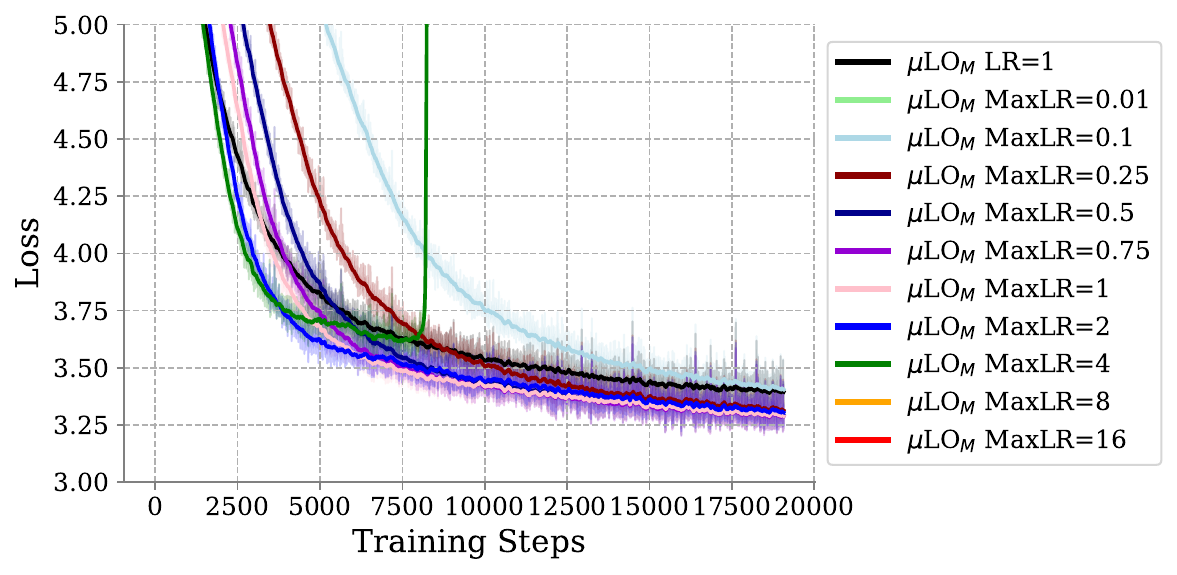}}\\
    \subfloat[$\mu$LO LR Tuning Ablation]{\includegraphics[width=0.7\linewidth]{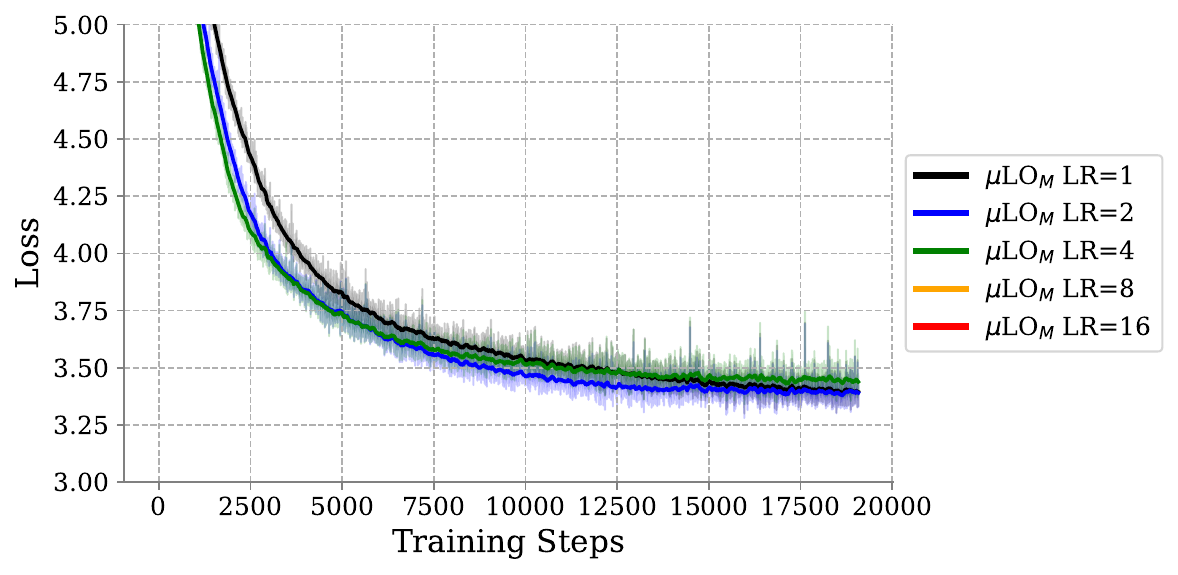}}
    \caption{\textbf{GPT2 Pre-training VeLO ablations: decoupled weight decay, LR-tuning, and cosine annealing.} We pre-train decoder-only transformers of different sizes on a causal language modeling objective. We estimate gradients from $512$ sequences of length $1024$, resulting in a $\sim 0.5$ M token batch size per step. We train all models for $10B$ tokens of FineWeb-EDU data. }
    \label{apdx:fig:gpt-mulo-wd-lr}
\end{figure*}

The language modeling experiments, presented in Figures \ref{apdx:fig:gpt-velo-wd-lr} and \ref{apdx:fig:gpt-mulo-wd-lr}, demonstrate more pronounced sensitivity to weight decay and scheduling compared to vision tasks. VeLO shows measurable performance improvements when augmented with weight decay and scheduling for specific parameter configurations. Similarly, MuLO benefits substantially from both weight decay regularization and cosine scheduling in the language modeling domain, suggesting that learned optimizers may benefit from scheduling and weight decay.\\

\textbf{Architectural Configuration Effects}

Figure \ref{apdx:fig:gpt-qkv} presents an additional ablation examining the impact of separate versus concatenated query-key-value (QKV) matrices on the optimizee's training loss. We observe that separating the QKV matrices in attention blocks improves performance across VeLO and $\mu$LO$_M$.

\begin{figure*}[h]
    \centering
    \subfloat[VeLO]{\includegraphics[width=0.5\linewidth]{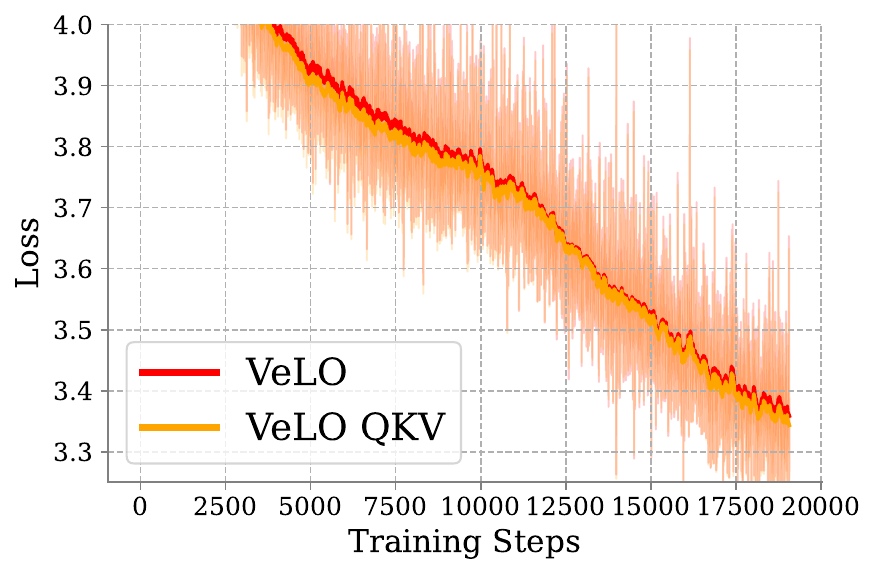}}
    \subfloat[$\mu$LO]{\includegraphics[width=0.5\linewidth]{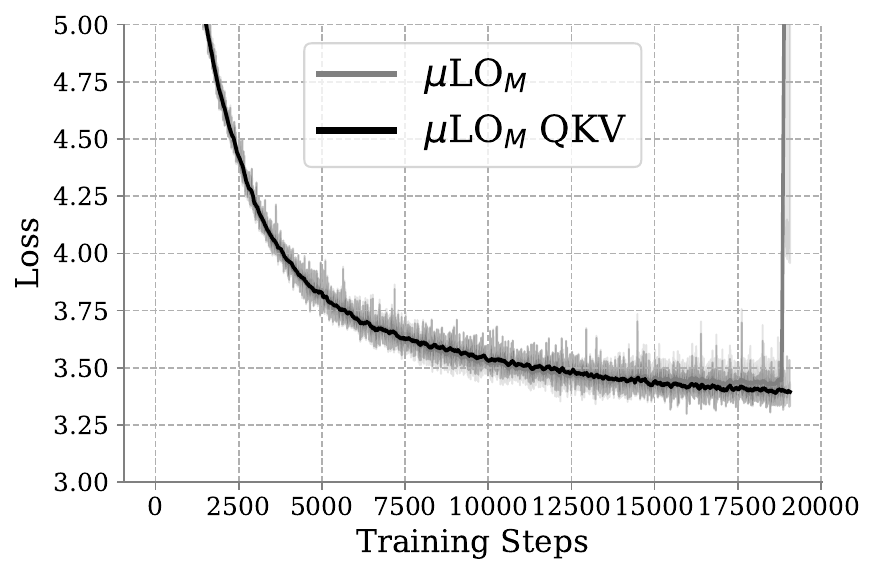}}

    \caption{\textbf{Separate QKV weight ablation.} We ablate using concatenated or separate QKV matrices in attention layers. From the perspective of the learned optimizer, these two settings will be treated differently. We observe that performance improves for both $\mu$LO$_M$ and VeLO when QKV matrices are separated.}
    \label{apdx:fig:gpt-qkv}
\end{figure*}
\clearpage
\section{Hyperparameters for Experiments}\label{apdx:sec:hparams}

\begin{table*}[ht]
    \centering
    \begin{minipage}[t]{0.48\textwidth}
    \caption{\textbf{Vision Transformer Training Hyperparameters.} }
    \label{table:vit-hyperparameters}
    \centering
    \begin{tabular}{ll}
    \toprule
    Description & Value\\\midrule
    \multicolumn{2}{l}{\textbf{Model Architecture}}\\
    Model & ViT-Base/16 \\
    Image Size & $224 \times 224$ \\
    Patch Size & $16 \times 16$ \\
    \multicolumn{2}{l}{\textbf{Training Configuration}}\\
    Batch Size & $4096$ \\
    Training Epochs & $480$ \\
    Optimizer (for baseline) & Adam \\
    Learning Rate ($\eta$) & $4 \times 10^{-3}$ \\
    LR Scheduler & Cosine \\
    Warmup Epochs & $32$ \\
    Weight Decay & $0.03$ \\
    Gradient Clipping & $1.0$ \\
    \multicolumn{2}{l}{\textbf{Data Augmentation}}\\
    Crop Percentage & $0.95$ \\
    Random Horizontal Flip & $0.5$ \\
    Mixup & $0.1$ \\
    CutMix & $1.0$ \\
    AutoAugment & rand-m7-mstd0.5 \\
    \multicolumn{2}{l}{\textbf{Regularization}}\\
    Dropout Rate & $0.1$ \\
    Stochastic Depth & $0.1$ \\
    \bottomrule
    \end{tabular}
    \end{minipage}
    \hfill
    \begin{minipage}[t]{0.48\textwidth}
    \centering
    \vspace{2pt}
    \caption{\textbf{Vision Transformer Model Variants Architecture.} }
    \label{table:vit-model-variants}
    \begin{tabular}{ll}
    \toprule
    Description & Value\\\midrule
    \multicolumn{2}{l}{\textbf{ViT-Small/16}}\\
    Parameters & $22.05$M \\
    Hidden Dimension ($d_{\text{model}}$) & $384$ \\
    MLP Hidden Dimension & $1536$ \\
    Number of Heads & $6$ \\
    Number of Layers & $12$ \\
    Patch Embedding Dim & $384$ \\
    \multicolumn{2}{l}{\textbf{ViT-Base/16}}\\
    Parameters & $86.57$M \\
    Hidden Dimension ($d_{\text{model}}$) & $768$ \\
    MLP Hidden Dimension & $3072$ \\
    Number of Heads & $12$ \\
    Number of Layers & $12$ \\
    Patch Embedding Dim & $768$ \\
    \multicolumn{2}{l}{\textbf{ViT-Large/16}}\\
    Parameters & $307.40$M \\
    Hidden Dimension ($d_{\text{model}}$) & $1024$ \\
    MLP Hidden Dimension & $4096$ \\
    Number of Heads & $16$ \\
    Number of Layers & $24$ \\
    Patch Embedding Dim & $1024$ \\
    \bottomrule
    \end{tabular}
    \end{minipage}
\end{table*}

\begin{table*}[ht]
    \centering
    \begin{minipage}[t]{0.48\textwidth}
    \caption{\textbf{GPT-2 Training Hyperparameters.} }
    \label{appen:table:gpt2-hyperparameters}
    \centering
    \begin{tabular}{ll}
    \toprule
    Description & Value\\\midrule
    \multicolumn{2}{l}{\textbf{Model Configuration}}\\
    Base Model & GPT-2 \\
    Architecture Type & Decoder-only Transformer \\
    Attention Mechanism & separate\_kqv \\
    Sequence Length ($T$) & $1024$ \\
    Vocabulary Size & $50257$ \\
    \multicolumn{2}{l}{\textbf{Training Configuration}}\\
    Batch Size & $512$ \\
    Training Iterations & $19073$ \\
    Optimizer (for baseline) & Adam \\
    Learning Rate Scheduler (for baseline) & Cosine \\
    Warmup Iterations & $381$ \\
    \multicolumn{2}{l}{\textbf{Regularization}}\\
    Attention Dropout & $0.1$ \\
    Embedding Dropout & $0.1$ \\
    Residual Dropout & $0.1$ \\
    \multicolumn{2}{l}{\textbf{Initialization}}\\
    Muon Initialization & Enabled \\
    Weight Initialization & Standard GPT-2 \\
    \bottomrule
    \end{tabular}
    \end{minipage}
    \hfill
    \begin{minipage}[t]{0.48\textwidth}
    \centering
    \vspace{2pt}
    \caption{\textbf{GPT-2 Model Variants Architecture.} }
    \label{table:gpt2-model-variants}
    \begin{tabular}{ll}
    \toprule
    Description & Value\\\midrule
    \multicolumn{2}{l}{\textbf{GPT-2 Small}}\\
    Parameters & $\sim36$M \\
    Hidden Dimension ($d_{\text{model}}$) & $768$ \\
    Number of Heads & $12$ \\
    Number of Layers & $12$ \\
    Head Dimension & $64$ \\
    FFN Hidden Dimension & $3072$ \\
    \multicolumn{2}{l}{\textbf{GPT-2 Medium}}\\
    Parameters & $\sim345$M \\
    Hidden Dimension ($d_{\text{model}}$) & $1024$ \\
    Number of Heads & $16$ \\
    Number of Layers & $24$ \\
    Head Dimension & $64$ \\
    FFN Hidden Dimension & $4096$ \\
    \multicolumn{2}{l}{\textbf{GPT-2 Large}}\\
    Parameters & $\sim762$M \\
    Hidden Dimension ($d_{\text{model}}$) & $2048$ \\
    Number of Heads & $32$ \\
    Number of Layers & $16$ \\
    Head Dimension & $64$ \\
    FFN Hidden Dimension & $8192$ \\
    \bottomrule
    \end{tabular}
    \end{minipage}
\end{table*}
\clearpage
\section{Validation of PyLO Implementation Against Original JAX Codebase}

To validate the correctness of our implementation of the popular learned optimizers in PyLO, we conducted a direct comparison with the original JAX implementation from Google's learned optimization\citep{metz2022practical} codebase. We evaluated both $\mu$LO\citep{therien2024mulo} and VeLO\citep{velo} algorithms on a simplified image classification benchmark using ImageNet resized to 64×64 pixels with a 3-layer MLP architecture (width 128). The comparison spans the first 5,000 training steps to assess implementation accuracy in a practical and economical setup.

\begin{figure}[H]
    \centering
    \subfloat[$\mu$LO implementation comparison\label{fig:jax_pylo_mulo}]{
        \includegraphics[width=0.48\textwidth]{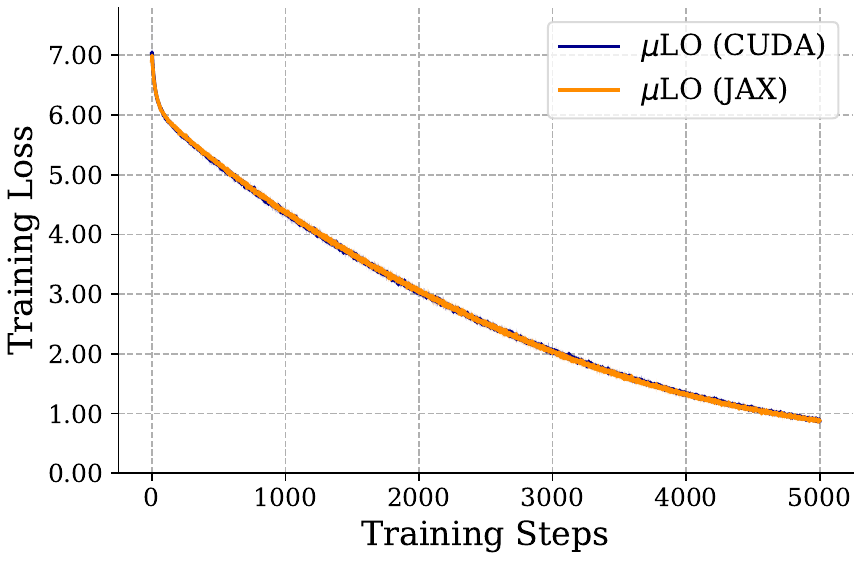}
    }
    \hfill
    \subfloat[VeLO implementation comparison\label{fig:jax_pylo_velo}]{
        \includegraphics[width=0.48\textwidth]{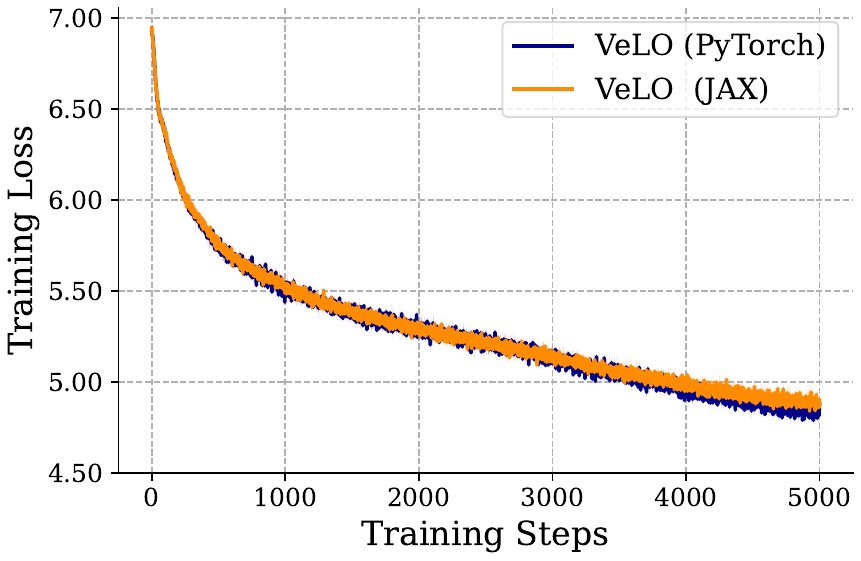}
    }
    \caption{Training curve comparison between original JAX implementation and PyLO library for $\mu$LO and VeLO optimizers on ImageNet 64×64 classification task using a 3-layer MLP (width 128). Both implementations show nearly identical convergence behavior over 5,000 training steps, validating the correctness of our implementation.}
    \label{fig:jax_pylo_comparison}
\end{figure}

The results demonstrate strong agreement between the two implementations across both optimizers. As shown in Figure~\ref{fig:jax_pylo_comparison}, the training curves exhibit nearly identical convergence patterns across all 5 different runs. Minor variations arise from inherent differences in how PyTorch and JAX compute gradients. Accounting for this, we confirm that our PyTorch implementation faithfully reproduces the behavior of the original JAX code. We further plan to compare both implementations across a range of models and tasks. 

\clearpage
\section{Extended per time step results}
\label{apdx:full-timing}

\begin{table*}[h]
\begin{adjustbox}{width=\textwidth,center}
\begin{tabular}{@{}l l c c c c c c c@{}}
\toprule
\textbf{Model} & \textbf{Optimizer} & \textbf{Batch Size} & \textbf{Samples/sec} & \textbf{Step time (ms)} & \textbf{Fwd (ms)} & \textbf{Bwd (ms)} & \textbf{Opt step (ms)} & \textbf{Params (M)} \\ \midrule

\multirow{6}{*}{ViT-B-16} 
  & AdamW                     & \multirow{6}{*}{32} & 524.09 & 60.55  & \multirow{6}{*}{17.52} & \multirow{6}{*}{38.13} & 4.90  & \multirow{6}{*}{86.57} \\
  & Adafactor                 &                     & 425.51 & 74.69  &                        &                        & 18.99 &                        \\
  & \texttt{small\_fc\_lopt} (naive) &              & 39.36  & 812.51 &                        &                        & 756.80&                        \\
  & \texttt{small\_fc\_lopt} (CUDA)  &              & 205.59 & 155.16 &                        &                        & 99.59 &                        \\
  & VeLO (naive)     &                     & 49.73  & 644.45 &                        &                        & 588.80&                        \\
  & VeLO (CUDA)      &                     & 191.18 & 168.61 &                        &                        & 112.97&                        \\ \midrule

\multirow{6}{*}{ViT-S-16} 
  & AdamW                     & \multirow{6}{*}{32} & 1,116.12 & 28.17 & \multirow{6}{*}{7.65}  & \multirow{6}{*}{18.27} & 2.25  & \multirow{6}{*}{22.05} \\
  & Adafactor                 &                     & 711.97  & 44.47 &                        &                        & 18.53 &                        \\
  & \texttt{small\_fc\_lopt} (naive) &              & 109.03  & 293.01&                        &                        & 266.90&                        \\
  & \texttt{small\_fc\_lopt} (CUDA)  &              & 382.73  & 83.28 &                        &                        & 57.45 &                        \\
  & VeLO (naive)     &                     & 94.84   & 337.68&                        &                        & 311.77&                        \\
  & VeLO (CUDA)      &                     & 242.96  & 132.49&                        &                        & 106.57&                        \\ \midrule

\multirow{6}{*}{ViT-L-16} 
  & AdamW                     & \multirow{6}{*}{32} & 175.95 & 181.36 & \multirow{6}{*}{52.18} & \multirow{6}{*}{113.89}& 15.28 & \multirow{6}{*}{304.33}\\
  & Adafactor                 &                     & 156.64 & 203.34 &                        &                        & 37.23 &                        \\
  & \texttt{small\_fc\_lopt} (naive) &              & 11.64  & 2,748.52&                       &                        & 2,582.30&                      \\
  & \texttt{small\_fc\_lopt} (CUDA)  &              & 70.69  & 451.69&                        &                        & 285.80&                        \\
  & VeLO (naive)     &                     & 15.41  & 2,077.28&                       &                        & 1,911.20&                      \\
  & VeLO (CUDA)      &                     & 76.74  & 418.29&                        &                        & 252.21&                        \\ \midrule

\multirow{6}{*}{GPT2-125 M} 
  & AdamW                     & \multirow{6}{*}{4}  & 17.80  & 224.72 & \multirow{6}{*}{73.20} & \multirow{6}{*}{144.24}& 7.29  & \multirow{6}{*}{125.26}\\
  & Adafactor                 &                     & 17.02  & 235.07 &                        &                        & 17.76 &                        \\
  & \texttt{small\_fc\_lopt} (naive) &              & 3.38   & 1,182.33&                       &                        & 964.89&                        \\
  & \texttt{small\_fc\_lopt} (CUDA)  &              & 11.71  & 341.46&                        &                        & 124.09&                        \\
  & VeLO (naive)     &                     & 3.48   & 1,148.81&                       &                        & 931.37&                        \\
  & VeLO (CUDA)      &                     & 11.25  & 355.49&                        &                        & 138.05&                        \\ \midrule

\multirow{6}{*}{GPT2-410 M} 
  & AdamW                     & \multirow{6}{*}{4}  & 6.56   & 609.83 & \multirow{6}{*}{197.41}& \multirow{6}{*}{392.29}& 20.12 & \multirow{6}{*}{355.92}\\
  & Adafactor                 &                     & 6.40   & 624.64 &                        &                        & 35.11 &                        \\
  & \texttt{small\_fc\_lopt} (naive) &              & 1.16   & 3,461.96&                       &                        & 2,872.17&                      \\
  & \texttt{small\_fc\_lopt} (CUDA)  &              & 4.40   & 908.69&                        &                        & 319.14&                        \\
  & VeLO (naive)     &                     & 1.34   & 2,976.82&                       &                        & 2,387.12&                      \\
  & VeLO (CUDA)      &                     & 4.58   & 873.41&                        &                        & 283.71&                        \\ \midrule

\multirow{6}{*}{GPT2-1B} 
  & AdamW                     & \multirow{6}{*}{4}  & 2.88   & 1,387.51& \multirow{6}{*}{442.49}& \multirow{6}{*}{896.81}& 48.21 & \multirow{6}{*}{912.95}\\
  & Adafactor                 &                     & 2.87   & 1,395.03&                       &                        & 54.08 &                        \\
  & \texttt{small\_fc\_lopt} (naive) &              & OOM    & OOM     &                       &                        & OOM   &                        \\
  & \texttt{small\_fc\_lopt} (CUDA)  &              & 1.98   & 2,025.26&                       &                        & 681.78&                        \\
  & VeLO (naive)     &                     & OOM    & OOM     &                       &                        & OOM   &                        \\
  & VeLO (CUDA)      &                     & 2.24   & 1,782.43&                       &                        & 443.13&                        \\
\bottomrule
\end{tabular}
\end{adjustbox}
\caption{
\textbf{Per-step training breakdown (averaged over 40 steps).} 
We compare AdamW, Adafactor, \texttt{small\_fc\_lopt}, and the  VeLO optimizer in both naive (PyTorch) and CUDA-accelerated implementations. 
CUDA versions significantly reduce optimizer step time and memory overhead, enabling competitive throughput—especially for large models. 
}
\label{tab:full-timing}
\end{table*}

\newpage
\section{Features for learned optimizer input}
\label{apdx:features_learned_opt}
In the following section, we introduce the input features for \texttt{small\_fc\_lopt} and  \texttt{VeLO} in Tables~\ref{apdx:table:features-lopt} and~\ref{apdx:table:features-velo-adafac}, respectively.

\renewcommand\tabularxcolumn[1]{m{#1}} 
\begin{table*}[ht]
\small
\begin{center}
    \caption{\textbf{MLP input features for \texttt{small\_fc\_lopt}.} We replicated the table from~\cite{therien2024mulo} for the reader's convenience with only slight modifications; We also keep the same figure caption. All the coefficients, $\beta_i$, are learnable parameters adjusted during meta-optimization. All feature calculations and scalings are reported for a hidden weight matrix $\mW \in \R^{m\times n}$ in an optimizee network following our proposed $\mu$-parameterization. Here, $n$ is the width and $m=kn$ for some constant $k\in\R$. \textbf{Notation.} The table will use $\nabla_{t,i}$ or $\nabla_{t,j}$ to indicate the variable's dependence on time $t$ and coefficient $\beta_i$ or $\beta_j$, respectively. $(\nabla_{t,j})_{r,c}$ will designate indexing into row $r$ and column $c$ of the quantity $\nabla_{t,j}$.}
    \label{apdx:table:features-lopt}
    \begin{tabularx}{\textwidth}{llX|l}
        \toprule
        \multicolumn{1}{c}{\textbf{Type}} & \multicolumn{1}{c}{\textbf{\#}} & \multicolumn{1}{c|}{\textbf{Description}} & \multicolumn{1}{c}{\textbf{Accumulator Update/Equation}} \\
        \midrule
        \multirow{7}{*}{\textbf{Accumulators}} 
        &3&Momentum accumulators with coefficients $\beta_i, i\in\{1,2,3\}$.                         & $\mM_{t, i} = \beta_i \mM_{t-1, i} + (1 - \beta_i) \nabla_t $ \\[3mm]
        &1& Second moment accumulator with coefficient $\beta_4$.                                   & $\mV_t = \beta_4 \mV_{t-1} + (1 - \beta_4) \nabla_t^2 $ \\[3mm]
        &3& Adafactor row accumulator with coefficients $\beta_i, i\in\{5,6,7\}$.                    & $\vr_{t, i} = \beta_i \vr_{t-1, i} + (1 - \beta_i)~\texttt{row\_mean}(\nabla_t^2) $ \\[3mm]
        &3& Adafactor accumulator with coefficients $\beta_i, i\in\{5,6,7\}$.                        & $\vc_{t, i} = \beta_i \vc_{t-1, i} + (1 - \beta_i)~\texttt{col\_mean}(\nabla_t^2) $ \\\midrule
        \multirow{11}{*}{\parbox{2cm}{\textbf{Accumulator} \\ \textbf{~~~Features}}} 
        &3& Momentum values normalized by the square root of the second moment for $i\in\{5,6,7\}$.  & $\displaystyle\frac{\mM_{t,i}}{\sqrt{\mV_t}}$ \\[3mm]
        &1&The reciprocal square root of the second moment value.                                    & $\displaystyle\frac{1}{\sqrt{\mV}}$ \\[3mm]
        &6&The reciprocal square root of the Adafactor accumulators.                                 & $\displaystyle\frac{1}{\sqrt{\vr_{t,i}}} \textsc{  or  } \frac{1}{\sqrt{\vc_{t,i}}}$ \\[3mm]
        &3& Adafactor gradient features for $i\in\{5,6,7\}$.                                         & $\displaystyle\nabla_t \cdot \sqrt{\frac{\frac{1}{m}\sum_{h=1}^m(\vr_{t, i})_h}{ \vr_{t, i} \vc_{t, i}^T}}$ \\[3mm]
        &3& Adafactor momentum features for $i,j\in\{(5,1),(6,2),(7,3)\}$.                           & $\displaystyle\mM_{t,j} \cdot \sqrt{\frac{\frac{1}{m}\sum_{h=1}^m(\vr_{t, i})_h}{ \vr_{t, i} \vc_{t, i}^T}} $ \\\midrule
        \textbf{Time Features}&11& Time features for $x \in \{ 1, 3, 10, 30, 100, 300, 1000, 3000, 10^4, 3\cdot10^4, 10^5\}$. &$\tanh{\left( \frac{t}{x} \right)}$ \\\midrule
        \multirow{2}{*}{\textbf{Parameters}} 
        &1&Parameter value.                                    & $\mW_{t}$ \\
        &1&Gradient value.                                     & $\nabla_{t}$ \\\midrule
        \textbf{Total}&$39$&--&--\\
    \bottomrule
    \end{tabularx}
\end{center}
\end{table*}

\renewcommand\tabularxcolumn[1]{m{#1}} 
\begin{table*}[ht]
\small
\begin{center}
    \caption{\textbf{MLP input features for \texttt{VeLO}.} We replicated the table from~\cite{therien2024mulo} for the reader's convenience with only slight modifications; We also keep the same figure caption. All feature calculations and scalings are reported for a hidden weight matrix $\mW \in \R^{m\times n}$ in an optimizee network following our proposed $\mu$-parameterization. Here, $n$ is the width and $m=kn$ for some constant $k\in\R$. \textbf{Notation.} The table will use $\nabla_{t,i}$ or $\nabla_{t,j}$ to indicate the variable's dependence on time $t$ and coefficient $\beta_i$ or $\beta_j$, respectively. $(\nabla_{t,j})_{r,c}$ will designate indexing into row $r$ and column $c$ of the quantity $\nabla_{t,j}$.}
    \label{apdx:table:features-velo-adafac}
    \begin{tabularx}{\textwidth}{llX|l}
        \toprule
        \multicolumn{1}{c}{\textbf{Type}} & \multicolumn{1}{c}{\textbf{\#}} & \multicolumn{1}{c|}{\textbf{Description}} & \multicolumn{1}{c}{\textbf{Accumulator Update/Equation}} \\
        \midrule
        \multirow{7}{*}{\textbf{Accumulators}} 
        &3&Momentum accumulators with coefficients $\beta_i, i\in\{1,2,3\}$.                         & $\mM_{t, i} = \beta_i \mM_{t-1, i} + (1 - \beta_i) \nabla_t $ \\[3mm]
        &1& Second moment accumulator with coefficient $\beta_4$.                                   & $\mV_t = \beta_4 \mV_{t-1} + (1 - \beta_4) \nabla_t^2 $ \\[3mm]
        &3& Adafactor row accumulator with coefficients $\beta_i, i\in\{5,6,7\}$.                    & $\vr_{t, i} = \beta_i \vr_{t-1, i} + (1 - \beta_i)~\texttt{row\_mean}(\nabla_t^2) $ \\[3mm]
        &3& Adafactor accumulator with coefficients $\beta_i, i\in\{5,6,7\}$.                        & $\vc_{t, i} = \beta_i \vc_{t-1, i} + (1 - \beta_i)~\texttt{col\_mean}(\nabla_t^2) $ \\\midrule
        \multirow{11}{*}{\parbox{2cm}{\textbf{Accumulator} \\ \textbf{~~~Features}}} 
        &3& Momentum values normalized by the square root of the second moment for $i\in\{5,6,7\}$.  & $\displaystyle\frac{\mM_{t,i}}{\sqrt{\mV_t}}$ \\[3mm]
        &1&The reciprocal square root of the second moment value.                                    & $\displaystyle\frac{1}{\sqrt{\mV}}$ \\[3mm]
        &6&The reciprocal square root of the Adafactor accumulators.                                 & $\displaystyle\frac{1}{\sqrt{\vr_{t,i}}} \textsc{  or  } \frac{1}{\sqrt{\vc_{t,i}}}$ \\[3mm]
        &3& Adafactor gradient features for $i\in\{5,6,7\}$.                                         & $\displaystyle\nabla_t \cdot \sqrt{\frac{\frac{1}{m}\sum_{h=1}^m(\vr_{t, i})_h}{ \vr_{t, i} \vc_{t, i}^T}}$ \\[3mm]
        &3& Adafactor momentum features for $i,j\in\{(5,1),(6,2),(7,3)\}$.                           & $\displaystyle\mM_{t,j} \cdot \sqrt{\frac{\frac{1}{m}\sum_{h=1}^m(\vr_{t, i})_h}{ \vr_{t, i} \vc_{t, i}^T}} $ \\\midrule
        \multirow{3}{*}{\textbf{Parameters}} 
        &1&Parameter value.                                    & $\mW_{t}$ \\
        &1&Gradient value.                                     & $\nabla_{t}$ \\
        &1&Clipped gradient value.                             & $\texttt{clip}(\nabla_{t},-0.1,0.1)$ \\\midrule
        \textbf{Total}&$29$&--&--\\
    \bottomrule
    \end{tabularx}
\end{center}
\end{table*}

\clearpage

\section{\rev{Wall clock times for Language modeling and Vision transformers}}
\label{apdx:wall_clock_times}

\begin{figure*}[h]
    \centering
    \subfloat[GPT-2 Medium FineWeb]{\includegraphics[width=0.45\linewidth]{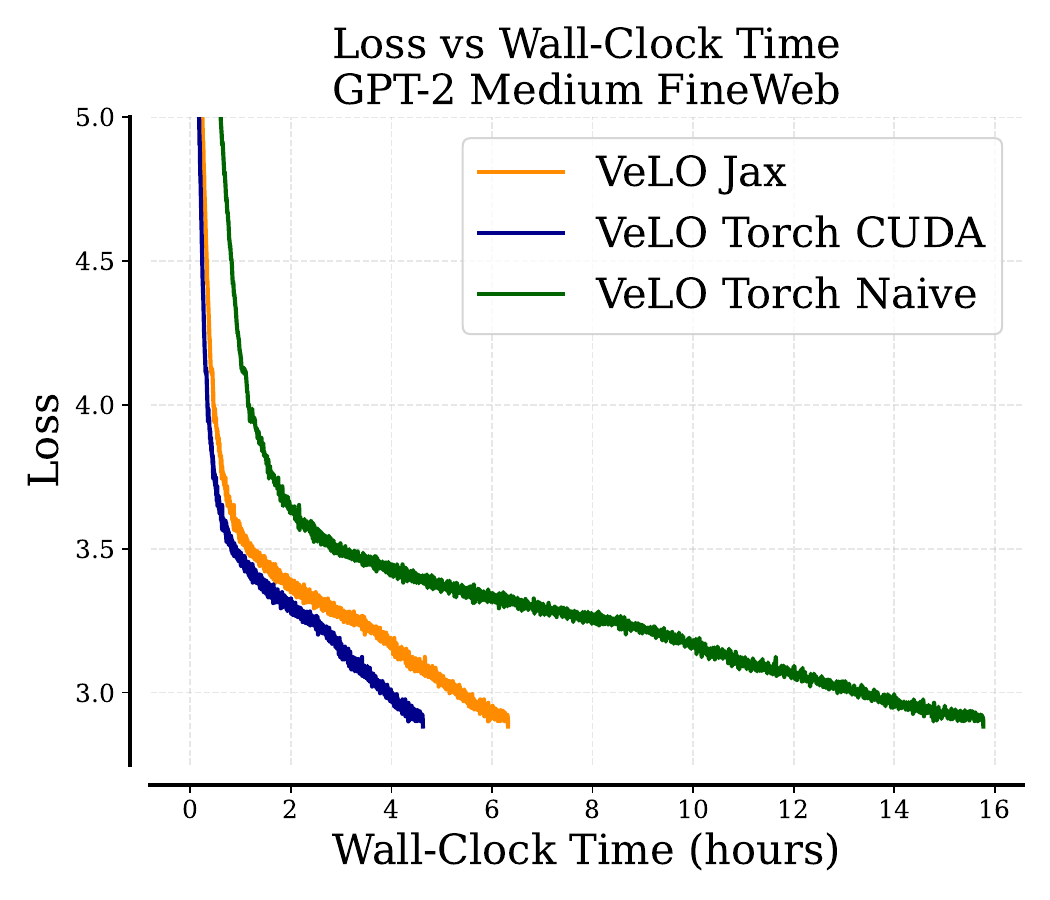}}
    \subfloat[ViT/B ImageNet]{\includegraphics[width=0.45\linewidth]{fig/velo_vs_time.pdf}}
\vspace{-10pt}
    \caption{\textbf{Wall-clock training time across different implementations of VeLO for (a) language and (b) vision tasks.} We observe that our CUDA kernel significantly accelerates training, leading to substantially improved wall-clock training times over the Jax and Naive implementations.  }
    \label{apdx:fig:gpt-qkv}
\end{figure*}

\clearpage
\section{\rev{Kernel name descriptions}}
\label{apdx:kernel_names}
  \begin{table}[h]                                                                                                                                                                                                                
  \centering                                                                                                                                                                                                                      
  \caption{CUDA Kernel Name Legend}                                                                                                                                                                                               
  \label{tab:kernel_legend}                                                                                                                                                                                                       
  \begin{tabular}{ll}                                                                                                                                                                                                             
  \toprule                                                                                                                                                                                                                        
  \textbf{Kernel Name} & \textbf{Description} \\                  
  \midrule                                                                                                                                                                                                                        
  \texttt{GEMM} & General Matrix Multiply (\texttt{ampere\_sgemm}), single-precision dense matmul \\
  \texttt{GEMM\_Reduce} & Split-K reduction kernel for tiled GEMM accumulation \\                                                                                                                                                 
  \texttt{GEMV} & General Matrix-Vector multiply (\texttt{gemvx::kernel}) \\                                                                                                                                                      
  \texttt{GEMV\_Batched} & Batched GEMV via cuBLAS (\texttt{cublasGemvParamsEx}) \\                                                                                                                                               
  \midrule                                                                                                                                                                                                                        
  \texttt{MSE} & Mean Squared Error forward kernel \\                                                                                                                                                                             
  \texttt{MSE\_Backward} & Gradient kernel for MSE loss \\                                                                                                                                                                        
  \texttt{Mean\_Reduce} & Reduction kernel computing the mean (\texttt{MeanOps}) \\
  \texttt{Reduce\_Max} & Reduction kernel computing the maximum (\texttt{MaxOps}) \\                                                                                                                                              
  \texttt{Reduce\_MaxNaN} & Reduction kernel computing the maximum, NaN-aware (\texttt{MaxNanFunctor}) \\                                                                                                                         
  \texttt{Reduce\_MaxNaN\_512} & Same as above with 512-thread block configuration \\                                                                                                                                             
  \texttt{Reduce\_Max\_128} & Max reduction with 128-thread block configuration \\                                                                                                                                                
  \texttt{Reduce\_1Block} & Single-block reduction kernel \\                                                                                                                                                                      
  \midrule                                                                                                                                                                                                                        
  \texttt{Add} & Elementwise addition \\                                                                                                                                                                                          
  \texttt{Multiply} & Elementwise multiplication \\               
  \texttt{Divide} & Elementwise division \\                                                                                                                                                                                       
  \texttt{Sqrt} & Elementwise square root \\                                                                                                                                                                                      
  \texttt{Pow} & Elementwise power \\                                                                                                                                                                                             
  \texttt{Lerp} & Elementwise linear interpolation \\                                                                                                                                                                             
  \texttt{Fill} & Elementwise fill with constant (\texttt{FillFunctor}) \\                                                                                                                                                        
  \texttt{Elementwise} & Generic elementwise kernel (unclassified) \\                                                                                                                                                             
  \midrule                                                                                                                                                                                                                        
  \texttt{Compute\_Squared\_Average\_LO} & Computes squared gradient averages for the learned optimizer \\                                                                                                                        
  \texttt{Apply\_LO} & Applies the learned optimizer parameter update \\                                                                                                                                                          
  \midrule                                                                                                                                                                                                                        
  \texttt{CatArray} & Batched concatenation along a tensor dimension \\                                                                                                                                                           
  \texttt{CatArray\_1D} & Batched concatenation, 1-D tensors \\                                                                                                                                                                   
  \texttt{CatArray\_2D} & Batched concatenation, 2-D tensors \\                                                                                                                                                                   
  \texttt{CatArray\_3D} & Batched concatenation, 3-D tensors \\                                                                                                                                                                   
  \bottomrule                                                                                                                                                                                                                     
  \end{tabular}                                                                                                                                                                                                                   
  \end{table}

\end{document}